\documentclass[10pt,twocolumn,letterpaper]{article}

\usepackage[utf8]{inputenc}  
\usepackage[T1]{fontenc}
\usepackage{iccv}
\usepackage{times}
\usepackage{epsfig}
\usepackage{graphicx}
\usepackage{amsmath}
\usepackage{amssymb}
\usepackage[table]{xcolor}
\usepackage[caption=false,font=footnotesize]{subfig}
\usepackage{booktabs}
\usepackage{multirow}
\usepackage{setspace}
\usepackage{stackengine}
\usepackage{microtype}
\usepackage{textcomp}
\usepackage{anyfontsize}
\usepackage{psfrag}
\usepackage{adjustbox}

\usepackage[pagebackref=true,breaklinks=true,letterpaper=true,colorlinks,bookmarks=false]{hyperref}

\iccvfinalcopy %

\def\iccvPaperID{6981} %

\ificcvfinal\fi
\begin{document}

\title{Physics-Based Rendering for Improving Robustness to Rain}

\author{\hspace{-0.5em}Shirsendu Sukanta Halder\\
\hspace{-0.5em}Inria, Paris, France\\
\hspace{-0.5em}{\tt\small shalder@cs.iitr.ac.in}
\and
Jean-François Lalonde\\
Université Laval, Québec, Canada\\
{\tt\small jflalonde@gel.ulaval.ca}
\and
\hspace{-0.5em}Raoul de Charette\\
\hspace{-0.5em}Inria, Paris, France\\
\hspace{-0.5em}{\tt\small raoul.de-charette@inria.fr}\\
\hspace{-30em}{\small\url{https://team.inria.fr/rits/computer-vision/weather-augment/}}\vspace{-1em}
}

\maketitle

\begin{abstract}
To improve the robustness to rain, we present a physically-based rain rendering pipeline for realistically inserting rain into clear weather images. 
Our rendering relies on a physical particle simulator, an estimation of the scene lighting and an accurate rain photometric modeling to augment images with arbitrary amount of realistic rain or fog.
We validate our rendering with a user study, proving our rain is judged $40\%$ more realistic that state-of-the-art.
Using our generated weather augmented Kitti and Cityscapes dataset, we conduct a thorough evaluation of deep object detection and semantic segmentation algorithms and show that their performance decreases in degraded weather, on the order of 15\% for object detection and 60\% for semantic segmentation. 
Furthermore, we show refining existing networks with our augmented images improves the robustness of both object detection and semantic segmentation algorithms. We experiment on nuScenes and measure an improvement of 15\% for object detection and 35\% for semantic segmentation compared to original rainy performance. 
Augmented databases and code are available on the project page.
\end{abstract}

\section{Introduction}

\begin{figure}
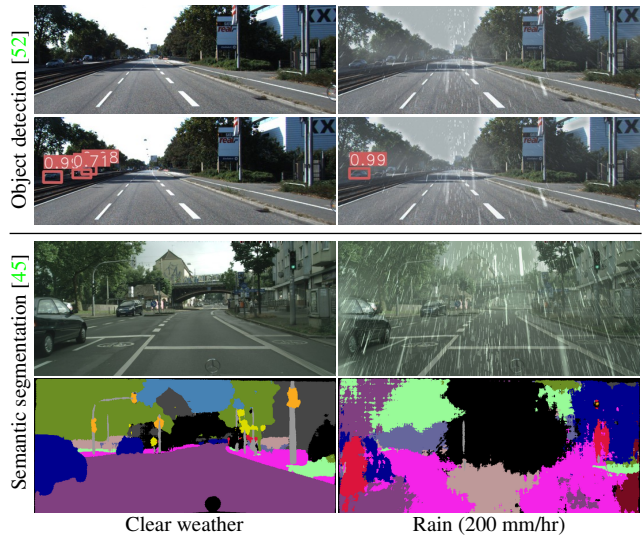

	\newcommand{\objectdetectframe}{000004}
	\newcommand{\semanticsframe}{bochum_000000_037039_leftImg8bit}
	
	\centering
	\footnotesize
	\setlength{\tabcolsep}{0.025cm}
	\renewcommand{\arraystretch}{0.5}
	\begin{tabular}{ccc}
		\multirow{2}{*}[1cm]{\rotatebox{90}{Object detection~\cite{yang2016exploit}}} &
		\adjincludegraphics[width=0.48\linewidth,trim={{0.1938\width} {0.01835\height} 0 {0.0055\width}},clip]{figures/results_new/qualitative/objDetection/rendering/\objectdetectframe_clear_lr.jpg} &
		\adjincludegraphics[width=0.48\linewidth,trim={{0.1938\width} 0 0 0},clip]{figures/results_new/qualitative/objDetection/rendering/\objectdetectframe_200mm_lr.jpg}\\
		& 
		\adjincludegraphics[width=0.48\linewidth,trim={{0.1938\width} 0 0 0},clip]{figures/results_new/qualitative/objDetection/MxRCNN/\objectdetectframe_MxRCNN_clear_lr.jpg} &
		\adjincludegraphics[width=0.48\linewidth,trim={{0.1938\width} 0 0 0},clip]{figures/results_new/qualitative/objDetection/MxRCNN/\objectdetectframe_MxRCNN_200mm_lr.jpg}\\
		\midrule
		\multirow{2}{*}[1.5cm]{\rotatebox{90}{Semantic segmentation~\cite{mehta2018espnet}}} &
		\adjincludegraphics[width=0.48\linewidth,trim={0 {0.110728\height} 0 0},clip]{figures/results_new/qualitative/rendering/cityscapes/clear/\semanticsframe_lr.jpg} &
		\adjincludegraphics[width=0.48\linewidth,trim={0 {0.110728\height} 0 0},clip]{figures/results_new/qualitative/semanticSeg/rendering/\semanticsframe_200mm_lr.jpg}\\
		& 
		\adjincludegraphics[width=0.48\linewidth,trim={0 {0.110728\height} 0 0},clip]{figures/results_new/qualitative/semanticSeg/ESPNet/\semanticsframe_ESPNet_clear_lr.png} &
		\adjincludegraphics[width=0.48\linewidth,trim={0 {0.110728\height} 0 0},clip]{figures/results_new/qualitative/semanticSeg/ESPNet/\semanticsframe_ESPNet_200mm_lr.png}\\
		& 
		Clear weather &
		Rain (200~mm/hr) \\
	\end{tabular}
	\vspace{.25em}
	\caption{Our synthetic rain rendering framework allows for the evaluation of computer vision algorithms in these challenging bad weather scenarios. We render physically-based, realistic rain on images from the Kitti~\cite{Geiger2012CVPR} (rows 1-2) and Cityscapes~\cite{Cordts2016Cityscapes} (rows 3-4) datasets with object detection from mx-RCNN~\cite{yang2016exploit} (row 2) and semantic segmentation from ESPNet~\cite{mehta2018espnet} (last row). Both algorithms are quite significantly affected by rainy conditions.}
	\label{fig:overview}
	\vspace{-1em}
\end{figure}

A common assumption in computer vision is that light travels, unaltered, from the scene to the camera. In clear weather, this assumption holds: the atmosphere behaves like a transparent medium and transmits light with very little attenuation or scattering. However, inclement weather conditions such as rain fill the atmosphere with particles producing spatio-temporal artifacts such as attenuation or rain streaks. This creates noticeable changes to the appearance of images (see fig.~\ref{fig:overview}), thus creating additional challenges to computer vision algorithms who must be robust to these conditions. 

Most, if not all computer vision practitioners know that bad weather affect our algorithms. %
However, very few of us actually know \emph{how much} it affects them. Indeed, how can one know what the impact of, say, a rainfall rate of 100~mm/hour (a typical autumn shower) will have on the performance of an object detector? Our existing databases all contain images overwhelmingly captured under clear weather conditions. To quantify this effect, one would need a labeled object detection dataset, where all the images have been captured under 100~mm/hour rain! Needless to say, such a ``rain-calibrated'' dataset does not exist, and capturing one would be prohibitive.

In this paper, we propose a method to realistically augment existing image databases with rainy conditions. Our method relies on well-understood physical models to generate visually convincing results.
Our approach is the first to allow controlling the \emph{amount} of rain in order to generate arbitrary amounts, ranging from very light rain (5~mm/hour rainfall) to very heavy storms (300+~mm/hour). 
This key feature allows us to produce weather-augmented datasets, where the rainfall rate is known and calibrated. Subsequently, we augment two existing datasets (Kitti~\cite{Geiger2012CVPR} and Cityscapes~\cite{Cordts2016Cityscapes}) with rain, and evaluate the robustness of popular computer vision algorithms on these augmented databases. We also use the latter to refine algorithms using curriculum learning~\cite{bengio2009curriculum} which demonstrate improved robustness on real rainy scenarios.

As opposed to the recent style transfer approaches~\cite{johnson2016perceptual,gatys2016image}, which have demonstrated the ability to transfer weather~\cite{zhu2017unpaired}, we use physical and photometric models to render all raindrops individually. Indeed, while these GAN-based approaches create visually appealing images, there is no guarantee they respect the underlying physics of weather, and thus cannot be used to estimate the performance of vision systems. What is more, controlling the \emph{amount} of rain cannot easily be achieved with these techniques. 

\begin{figure}
\centering
\includegraphics[width=\linewidth]{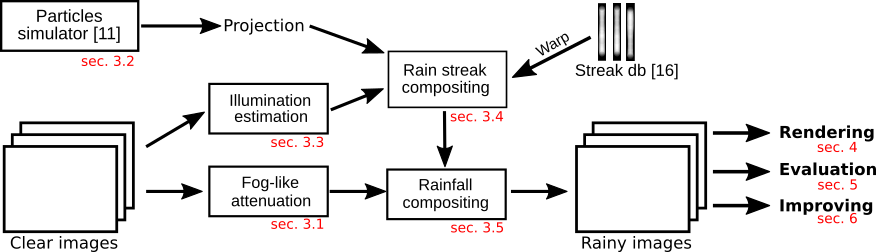}
\caption{Overview of our weather augmentation pipeline for synthetically adding realistic rain in existing image databases.}
\label{fig:system}
\vspace{-1em}
\end{figure}

In short, we make the following contributions. 
First, we present a practical, physically-based approach to render realistic rain in images (fig.~\ref{fig:system}). 
Second, we augment Kitti~\cite{Geiger2012CVPR} and Cityscapes~\cite{Cordts2016Cityscapes} datasets with rain. %
Third, we present a methodology for systematically evaluating the performance of 12 popular object detection and semantic segmentation algorithms. Our findings indicate that most algorithms decrease on the order of 15\% for object detection, and 60\% for semantic segmentation.
Finally, our augmented database is used to finetune object detection and segmentation architecture leading to significantly better robustness in real-world rainy/clear conditions.

\section{Related Work}
\label{sec:related-work}

\paragraph{Rain modeling} In their series of influential papers, Garg and Nayar provided a comprehensive overview of the appearance models required for understanding~\cite{garg2007vision} and synthesizing~\cite{garg2006photorealistic} realistic rain. In particular, they propose an image-based rain streak database~\cite{garg2006photorealistic} modeling the drop oscillations, which we exploit in our rendering framework. 
Other streak appearance models were proposed in \cite{weber2015multiscale,barnum2010analysis} using a frequency model. Realistic rendering was also obtained with ray-tracing \cite{rousseau2006realistic} or artistic-based techniques \cite{tatarchuk2006artist,creus2013r4} but on synthetic data as they requires complete 3D knowledge of the scene including accurate light estimation. 

\paragraph{Rain removal} Due to the problems it creates on computer vision algorithms, rain removal in images got a lot of attention initially focusing on photometric models~\cite{garg2004detection}. For this, several techniques have been proposed, ranging from frequency space analysis~\cite{barnum2010analysis} to deep networks~\cite{yang2017deep}. Sparse coding and layers priors were also important axes of research \cite{li2016rain,luo2015removing,chen2013generalized} due to their facility to encode streak patches.
More recently, Zhang et al.~\cite{zhang2017image} proposed to use conditional GANs for this task. Alternatively, camera parameters~\cite{garg2005does} or programmable light sources~\cite{de2012fast} can also be adjusted to limit the impact of rain on the image formation process. 
Additional proposal were made for the specific of raindrop removals on windows \cite{eigen2013restoring} or windshields \cite{halimeh2009raindrop}.

\paragraph{Weather databases} In computer vision, few images databases have precise labeled weather information. 
Of note for mobile robotics, the BDD100K~\cite{yu2018bdd100k} and the Oxford dataset~\cite{RobotCarDatasetIJRR} provides data recorded in various weathers including rain. 
Other stationary cameras datasets such as AMOS~\cite{jacobs07amos}, the transient attributes dataset~\cite{Laffont14}, the Webcam Clip Art dataset~\cite{lalonde2009webcam}, or the WILD dataset~\cite{narasimhan2002all} are sparsely labeled with weather information.
As of yet, there exists no dataset with systematically-recorded rainfall rates and object/scene labels. 
The closest systematic work in spirit \cite{johnson2016driving} evaluated the effect of simulating weather on vision but in a virtual game GTA.
Of particular relevance to our work, Sakaridis et al.~\cite{SDV18} propose a framework for rendering fog into images from the Cityscapes~\cite{Cordts2016Cityscapes} dataset. Their approach assumes a homogenous fog model, which is rendered from the depth estimated from stereo. Existing scene segmentation models and object detectors are then adapted to fog. 
In our work, we employ the similar idea of rendering realistic weather to existing images, but focus on rain rather than fog. 

\noindent{}We emphasize that rendering rain is significantly harder than fog, as it requires the accurate modeling of dynamics of rain drops and the radiometry of the resulting rain streaks as they get imaged by a camera. We also render fog, using a more realistic heterogeneous model.
\section{Rendering rain into images}
\label{sec:rendering.rain}

Rain is the result of moisture condensation at high altitude which creates raindrops (0.1-10~mm) falling at high speeds (up to 10~m/s)~\cite{marshall1948distribution,van1997numerical}. 
The intensity of rain is measured as the rain fall average over an hour. 
As a reference, moderate rain is roughly 10~mm/hr and heavy rain greater than 50~mm/hr. In non-tropical regions, extreme rain fall rates 200+ mm/hr usually occur during a short period of time, usually on the order of a few minutes only.

Because of their large size relative to the light wavelength, the interaction with the light is complex. As opposed to fog~\cite{SDV18}, rainy events cannot be modeled as a volume and each drop physics must be modeled separately. In this section, we describe our rain rendering pipeline which requires an image, its associated per-pixel depth, and a desired rainfall rate. From this information, rain is rendered and blended with the original image (see fig.~\ref{fig:overview}). 

We make the distinction between two types of raindrops, based on their imaging size. First, when they are too far away, the accumulation of drops image within the pixel cone attenuates the light in a fog-like manner. When they are closer to the camera and thus larger, falling drops produce motion blurred streaks. We render these two effects separately. 

\subsection{Fog-like rain}
\label{sec:rendering.fog-like}

We begin by rendering fog-like rain, which are the set of drops that are too far away and thus imaged on less than 1 pixel. In this case, a pixel may even be imaging a large number of drops, which causes optical attenuation~\cite{garg2007vision}.
In practice, most drops in a rainfall are actually imaged as fog-like rain\footnote{Assuming a stationary camera with Kitti calibration, we computed that only 1.24\% of the drops project on 1+ pixel in 50~mm/hr rain, and 0.7\% at 5~mm/hr. This is logical, as the heavier the rain, the higher the probability of having large drops.}, though their visual effect is less dominant.

We render volumetric attenuation using the model described in \cite{weber2015multiscale} where the per-pixel attenuation $I_\text{att}$ is expressed as the sum of the extinction $L_\text{ext}$ caused by the volume of rain and the airlight scattering $A_\text{in}$ that results of the environmental lighting. Using equations from \cite{weber2015multiscale} to model the attenuation image at pixel $\mathbf{x}$ we obtain
\begin{equation}
\label{eq:render-foglike}
I_\text{att}(\mathbf{x}) = I L_\text{ext}(\mathbf{x}) + A_\text{in}(\mathbf{x}) \,, \text{where}
\end{equation}
\begin{equation}
\begin{split}
L_\text{ext}(\mathbf{x}) &= e^{-0.312R^{0.67}d(\mathbf{x})}  \,, \text{and} \\
A_\text{in}(\mathbf{x})  &= \beta_\text{HG}(\theta) \bar{E}_\text{sun} (1 - L_\text{ext}(\mathbf{x}))  \,.
\end{split}
\end{equation}

Here, $R$ denotes the rain fall rate $R$ (mm/hr), $d(\mathbf{x})$ the pixel depth, $\beta_\text{HG}$ the standard Heynyey-Greenstein coefficient, and $\bar{E}_\text{sun}$ the average sun irradiance which we estimate from the image-radiance relation~\cite{horn1986robot}.

\subsection{Simulating the physics of raindrops}
\label{sec:raindrop-position}

We use the particles simulator of de Charette et al.~\cite{de2012fast}, which computes the position and dynamics of all raindrops for a given fallrate\footnote{The distribution and dynamics of drops varies on earth due to gravity and atmospheric conditions. We selected here the broadly used physical models recorded in Ottawa, Canada~\cite{marshall1948distribution,atlas1973doppler}.}. To reduce the algorithm complexity,
 only drops of 1~mm or more are individually simulated which is reasonable since smaller ones are unlikely to be visible. 
In particular, the simulator computes the position and dynamics (start and end points of streaks) of all the rain particles in both world and image space, and accounts for intrinsic and extrinsic calibration for image projection.
We calibrate the simulator as to generate particles that cover the entire field of view of the camera.

\subsection{Rendering the appearance of raindrops}
\label{sec:rwa_appearance}

Visually, raindrops act as small spherical lenses imaging a wide portion of the scene. While it is possible to synthesize the exact photometry of a drop with ray casting, this comes at very high processing cost and is virtually only possible in synthetic scenes where the geometry and surface materials are perfectly known~\cite{rousseau2006realistic}. What is more, drops oscillate as they fall, which creates further complications in modeling the light interaction.

Instead, we rely on the raindrop appearance database of Garg and Nayar~\cite{garg2007vision}, which contains the individual rain streaks radiance when imaged by a stationary camera. For each drop the streak database also models 10 oscillations due to the airflow, which accounts for much greater realism than Gaussian modeling~\cite{barnum2010analysis}. 

\subsubsection{Projecting rain streak in the image}

To render a raindrop, we first select a rain streak $S_j^k \in \mathcal{S}$ from the streak database $\mathcal{S}$ of Garg and Nayar~\cite{garg2006photorealistic}, which contains $j=20$ different streaks with $k=10$ different oscillations stored in image format. 
To select the best rain streak for a particular drop, we pick the model $j$ that best matches the final drop dimensions (computed from the output of the physical simulator), and randomly select an oscillation $k$.

The selected rain streak $S$ (indices are dropped for clarity in notation) is subsequently warped to make it match the drop dynamics from the physical simulator: 
\begin{equation}
S' = \mathcal{H}(S) \,,
\label{eq:drop_warp}
\end{equation}
where $\mathcal{H}(\cdot)$ is the homography computed from the start and end points in image space given by the physical simulator and the corresponding points in the trimmed streak image.

\begin{figure}
	\centering
	\footnotesize
	\subfloat[Drop FOV]{\includegraphics[width=0.24\linewidth]{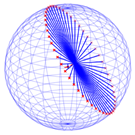}}\hspace{0.05\linewidth}
	\subfloat[Projection on the environment map]{\includegraphics[width=0.7\linewidth]{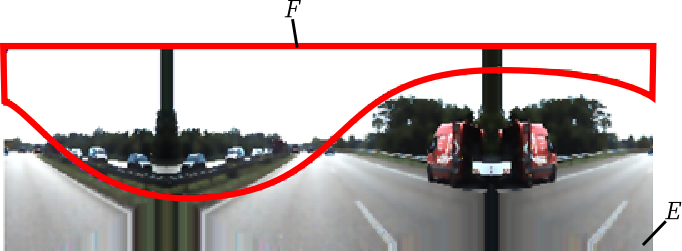}}
	\vspace{.25em}
	\caption{To estimate the photometric radiance of each drop, we must integrate the lighting environment map over the $165^\circ$ drop field of view (a). For this, we first estimate the environment map $E$ from the current image using \cite{Cameron2005}, and compute $F$ the intersection of the drop field of view with the environment map (right).}
	\label{fig:drop_cam_fov}
\end{figure}

\subsubsection{Computing the photometry of rain streak}
\label{sec:photometry-rainstreak}

Computing the photometry of a rain streak from a single image is challenging because drops have a much larger field of view than common cameras ($165^\circ$ vs approx. 70--100$^\circ$). 
In other words, a drop refracts light that is not imaged by the camera, which means that, if we are to render a drop accurately, we must estimate the environment map (spherical lighting representation) around that drop. While this is physically infeasible to perform from a single image, we employ the method of \cite{Cameron2005} which approximates the environment map through a series of simple operations performed on the image itself. 

From the estimated environment and the 3D drop position provided by the physical simulator, we compute the intersection $F$ of the drop field of view with the environment map $E$, assuming a 10~m constant scene distance and accounting for the camera-to-drop direction.
The process is depicted in fig.~\ref{fig:drop_cam_fov}, and geometrical details are provided in supplementary. 

Note that geometrically exact drop field of view estimation requires location-dependent environment maps, centered on each drop. However, we consider the impact negligible since drops are relatively close to the camera center compared to the sphere radius used\footnote{We computed that, for Kitti, 98.7\% of the drops are within 4~m from the camera center in a 50~mm/hr rainfall rate. Therefore, computing location-dependent environment maps would not be significantly more accurate,  while being of very high processing cost.}. 

Following \cite{garg2007vision} which states that a drop refracts 94\% of its field of view radiance and reflects 6\% of the entire environment map radiance, we multiply the streak appearance with a per-channel weight: 
\begin{equation}
S' = S' (0.94 \bar{F} + 0.06 \bar{E}) \,,
\label{eq:photometry_rainstreak}
\end{equation}
where $\bar{F}$ is the mean of the intersection region $F$, and $\bar{E}$ is the mean of the environment map $E$. Here, solid angles $\omega$ of the latitude-longitude representation are taken into account when computing the mean, i.e.: $\bar{F} = \sum_{i \in F} F(i) \omega(i)$.

\subsection{Compositing a single rain streak on the image}
\label{sec:rendering.streak-compositing}

Now that the streak position and photometry were determined from the physical simulator and the environment map respectively, we can composite it onto the original image. 
First, to account for the camera depth of field, we apply a defocus effect following~\cite{potmesil1981lens}, convolving the streak image $S'$ with the circle of confusion $C$\footnote{The circle of confusion $C$ of an object at distance $d$, is defined as: $C = \frac{(d - f_{\text{p}}) f^2}{d(f_{\text{p}} - f) f_{\text{N}}}$ with $f_\text{p}$ the focus plane, $f$ the focal and $f_\text{N}$ the lens f-number. $f$ and $f_\text{N}$ are from intrinsic calibration, and $f_\text{p}$ is set to 6~m.}, that is: $S' = S' * C$.

We then blend the rendered drop with the attenuated background image $I_{\text{att}}$, using the photometric blending model from \cite{garg2007vision}. Because the streak database and the image $I$ are likely to be imaged with different exposure, we need to correct the exposure so as it reflects the imaging system used in $I$.
Suppose $\mathbf{x}$ a pixel of the image $I$ and $\mathbf{x}'$ the overlapping coordinates in streak $S'$, the result of the blending is obtained with
\begin{equation}
I_{\text{rain}}(\mathbf{x}) = \frac{T - S'_{\alpha}(\mathbf{x'})\tau_1}{T}I_{\text{att}}(\mathbf{x}) + S'(\mathbf{x'})\frac{\tau_1}{\tau_0}\,,
\label{eq:rain_blending}
\end{equation}
where $S'_{\alpha}$ is the alpha channel\footnote{While \cite{garg2006photorealistic} does not provide an alpha channel, the latter is easily computed since drops were rendered on black background in a white ambient lighting.} of the rendered streak, $\tau_0 = \sqrt{10^{-3}} / 50$ is the time for which the drop remained on one pixel in the streak database, and $\tau_1$ the same measure according to our physical simulator. We refer to the supplementary materials for details.

\subsection{Compositing rainfall on the image}

The rendering of rainfall of arbitrary rates in an image is done in three main steps:
1)~the fog-like attenuated image $I_{\text{att}}$ is rendered (eq.~\ref{eq:render-foglike}), 
2)~the drops output by the physical simulator are rendered individually on the image (eq.~\ref{eq:rain_blending}),
3)~the global luminosity average of the rainy image denoted $I_{\text{rain}}$ is adjusted. 
While rainy events usually occur in cloudy weather which consequently decreases the scene radiance, a typical camera imaging system adjusts its exposure to restore the luminosity. Consequently, we adjust the global luminosity factor so as to restore the mean radiance, and preserves the relation $\bar{I} = \bar{I}_{\text{rain}}$, where the overbar denotes the intensity average.

\section{Validating rain appearance}
\label{sec:rwa_appearance_validation}

We now validate the appearance of our synthetic rain photometrically, visually and quantify the perceptual realism by comparing it to existing rain augmentation databases.

\begin{figure}
	\centering
	\footnotesize
	\setlength{\tabcolsep}{0.02cm}
	\renewcommand{\arraystretch}{0.5}
	\begin{tabular}{ccc}
		\multirow{1}{*}[1.35cm]{\rotatebox{90}{Ground truth}}&\includegraphics[width=0.31\linewidth, height=0.19\linewidth]{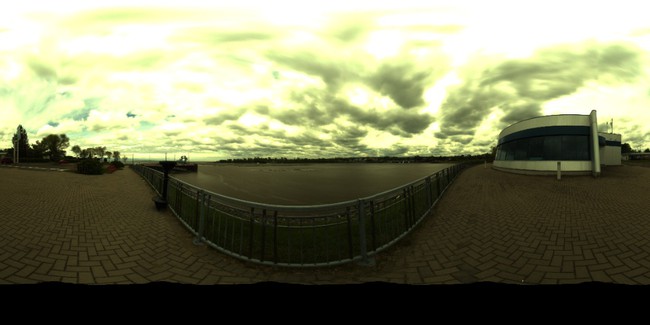}&\includegraphics[width=0.65\linewidth]{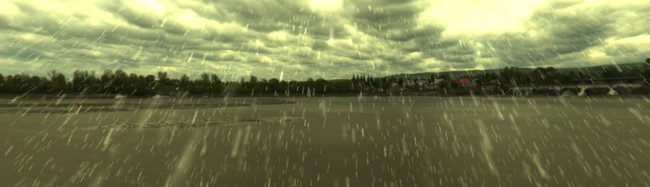}\\
		\multirow{1}{*}[0.85cm]{\rotatebox{90}{Ours}}&\includegraphics[width=0.31\linewidth, height=0.19\linewidth]{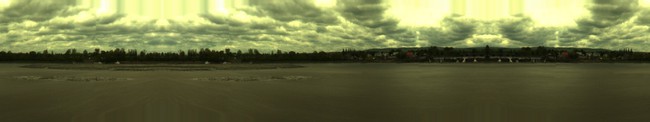}&\includegraphics[width=0.65\linewidth]{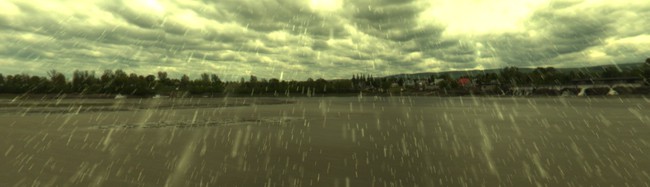}\vspace{0.05in}\\
		\multirow{1}{*}[1.35cm]{\rotatebox{90}{Ground truth}}&\includegraphics[width=0.31\linewidth, height=0.19\linewidth]{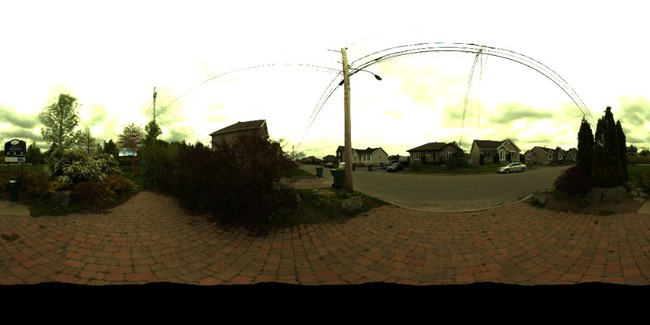}&\includegraphics[width=0.65\linewidth]{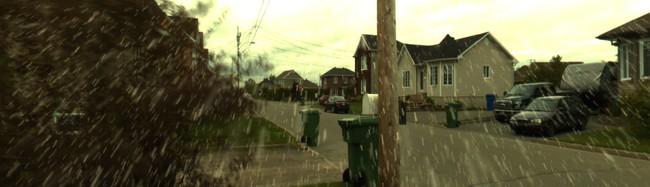}\\
		\multirow{1}{*}[0.85cm]{\rotatebox{90}{Ours}}&\includegraphics[width=0.31\linewidth, height=0.19\linewidth]{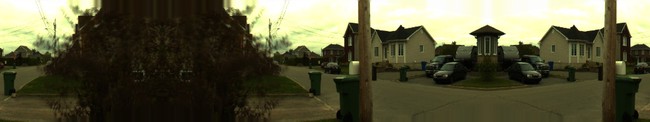}&\includegraphics[width=0.65\linewidth]{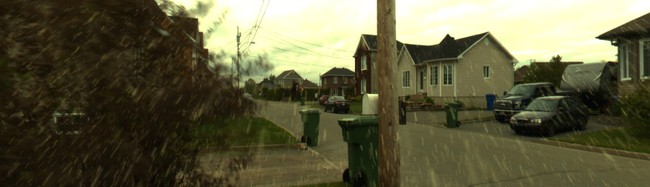}\\
		& Environment map & Our synthesized rain (50mm/hr) \\
	\end{tabular}
	\vspace{.25em}
	\caption{Comparison between rain rendering using ground truth illumination or our approximated environment map. 
	From HDR panoramas~\cite{holdgeoffroy-cvpr-19}, we first extract limited field of view crops to simulate the point of view of a regular camera. Then, 50mm/hr rain is rendered using either (rows 1, 3) the ground truth HDR environment map or (rows 2, 4) our environment estimation. The environments are shown as reference on the left. While our approximated environment maps differ from the ground truth, they are sufficient to generate visually similar rain in images.}
	\label{fig:pano_ours_comparison}
\end{figure}

\paragraph{Photometric validation.}
We first evaluate the impact of warping the background image to model the lighting environment around a drop.
For that matter, in fig.~\ref{fig:pano_ours_comparison} we compare rain rendered with our pipeline using the estimated environment map (sec.~\ref{sec:photometry-rainstreak}) with the ground truth illumination obtained from high dynamic range panoramas~\cite{holdgeoffroy-cvpr-19}. Overall, while this is clearly an approximation, we observe that rendering rain with this approximation closely match the results obtained with the ground truth lighting conditions. This is especially true when the scene is symmetrical (top image).

\begin{figure}
	\footnotesize
	\setlength{\tabcolsep}{0.025cm}
	\renewcommand{\arraystretch}{0.5}
	\begin{tabular}{ccc}
	 \multicolumn{3}{c}{\textbf{Rain photographs}}\\
	 &\adjincludegraphics[width=0.44\linewidth,trim={0 {0.21\height} 0 0},clip]{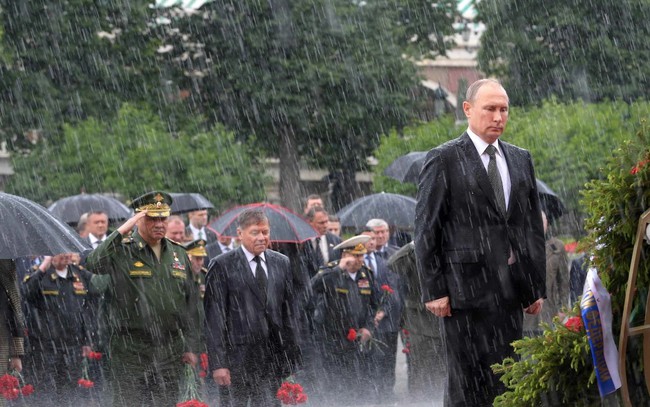}&\adjincludegraphics[width=0.47\linewidth,trim={0 {0.42\height} 0 0},clip]{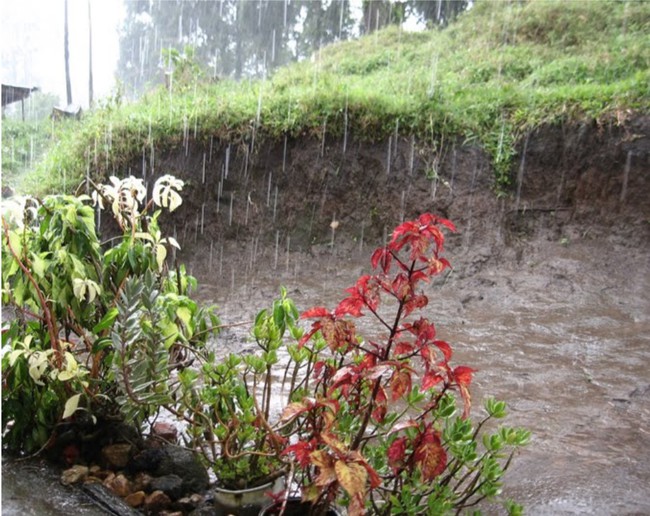}\\
	 \toprule
	 \multicolumn{3}{c}{\textbf{Our rain rendering}}\\
	\multirow{1}{*}[0.8cm]{\rotatebox{90}{Clear}}&\adjincludegraphics[width=0.47\linewidth,trim={{0.15225147637\width} 0 {0.13841043307\width} 0},clip]{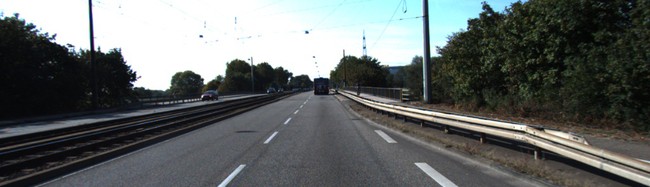}&\adjincludegraphics[width=0.47\linewidth,trim={0 {0.18823818897\height} 0 0},clip]{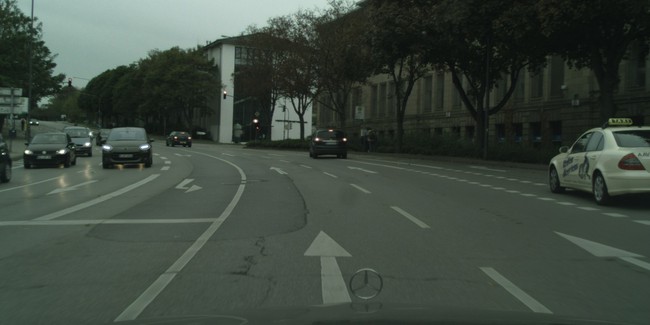}\\
	\multirow{1}{*}[1.2cm]{\rotatebox{90}{100mm/hr}}&\adjincludegraphics[width=0.47\linewidth,trim={{0.15225147637\width} 0 {0.13841043307\width} 0},clip]{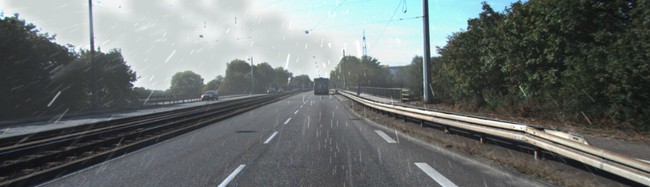}&\adjincludegraphics[width=0.47\linewidth,trim={0 {0.18823818897\height} 0 0},clip]{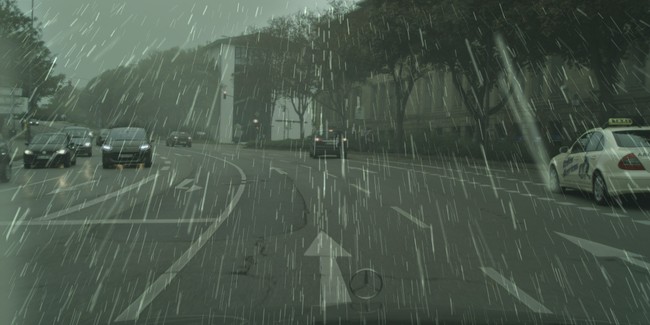}\\
	\multirow{1}{*}[1.2cm]{\rotatebox{90}{200mm/hr}}&\adjincludegraphics[width=0.47\linewidth,trim={{0.15225147637\width} 0 {0.13841043307\width} 0},clip]{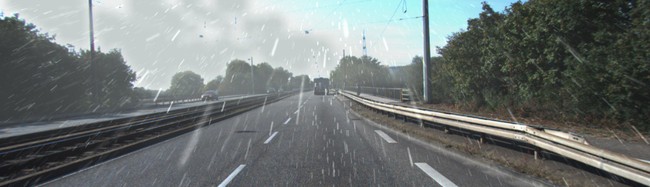}&\adjincludegraphics[width=0.47\linewidth,trim={0 {0.18823818897\height} 0 0},clip]{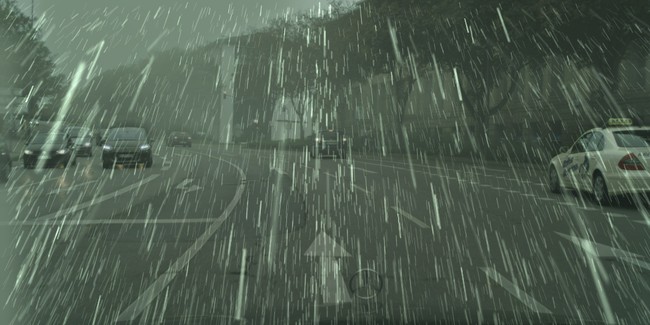}\\
	\end{tabular}
	\begin{tabular}{cccc}
	 \toprule
	 \multicolumn{4}{c}{\textbf{Other rain rendering}}\\ 	\multirow{1}{*}[1.3cm]{\rotatebox{90}{\textit{Unknown}}}&\adjincludegraphics[width=0.31\linewidth,trim={0 0 0 {0.03532268746\height}},clip]{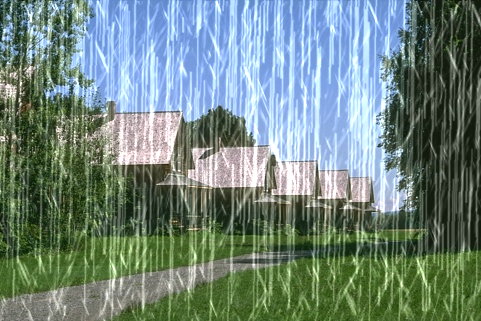}&\adjincludegraphics[width=0.31\linewidth,trim={0 0 0 {0.13\height}},clip]{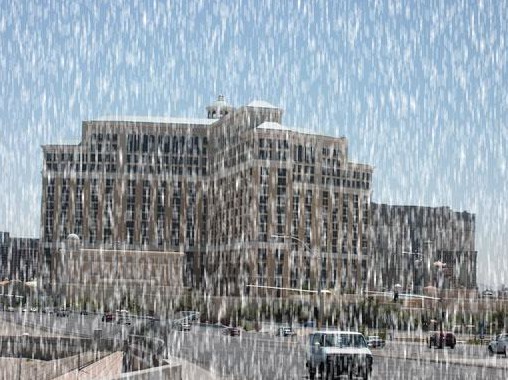}&\adjincludegraphics[width=0.31\linewidth,trim={0cm {0.08795508549\height} 0cm {0.25591017099\height}},clip]{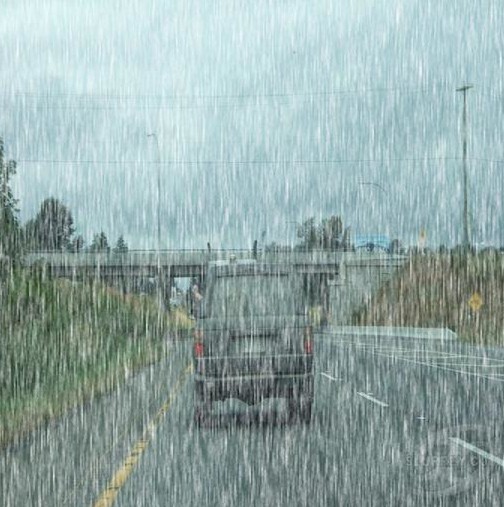}\\
	 & rain100H \cite{yang2017deep} & rain800 \cite{zhang2017image} & did-MDN \cite{zhang2018density}
	\end{tabular}
	\caption{Real photographs (source: web, \cite{luo2015removing}) showing heavy rain, sample output of our rain rendering and other recent rain rendering methods. Although rain appearance is highly camera-dependent \cite{garg2005does}, results show that both real photographs and our rain generation share volume attenuation and sparse visible streaks which correctly vary with the scene background. Opposed to the other rain renderings, our pipeline simulates physical rainfall (here, 100mm/hr and 200mm/hr) and valid particles photometry.}
	\label{fig:comparison-rain}
\end{figure}

\paragraph{Qualitative validation.}
Fig.~\ref{fig:comparison-rain} presents real photographs of heavy rain, and qualitative results of our rain rendering and representative results from 3 recent synthetic rain augmented databases~\cite{zhang2018density,yang2017deep,zhang2017image}. 
From our rendering, the streaks have consistent orientation with the camera motion and consistent photometry with background and depth (sec.~\ref{sec:eval-datasets-prepare} for details). 
As in the real photographs our streaks are sparse and only visible on darker background. 
Note also that the attenuation caused by the rain volume (i.e. fog-like rain) is visible where the scene depth is large (i.e. image center, sky) and that close streaks are accurately defocused. As opposed to other rain rendering, our pipeline simulates rain given real rainfall (mm/hr) whereas existing methods use arbitrary amount of rain that has no physical correspondence. 
This is important as we intend to study the effect of rain on computer vision.

\begin{figure}
	\centering
	\includegraphics[width=0.9\linewidth]{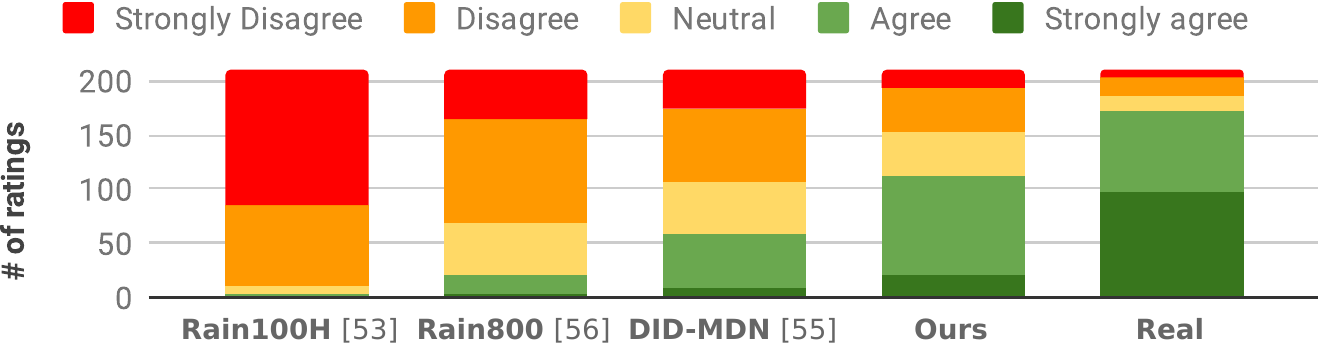}
	\caption{User study of rain realism. The $y$-axis displays ratings to the statement \textit{Rain in this image looks realistic}. Our rain is closer to real rain ratings method and outperforms all other methods.}
	\label{fig:userstudy}
\end{figure}

\paragraph{User study.}
We validate the perceptual quality of our synthesized rain through a Mean Opinion Score (MOS) user study with 35 participants which ages range from 19 to 46 (avg~25.9, std~6.7), with 40\% females. 
Users were asked to rate if \textit{rain looks realistic} on 30 randomly-selected images, using a 5-points Likert scale.
Results are reported in Fig.~\ref{fig:userstudy} against best images from \cite{yang2017deep,zhang2018density,zhang2017image} and real rain photography (6 images were shown for each method, in randomized order). 
Our rain is judged to be significantly more realistic than the state-of-the-art.
Converting ratings to the [0,~1] interval, the mean rain realism is 0.78 for real photos, 0.57 for ours and 0.41/0.31/0.12 for \cite{zhang2018density}/[\cite{zhang2017image}/\cite{yang2017deep}, respectively.


\begin{figure}
	\centering
	\subfloat[Rain]{\includegraphics[width=0.40\linewidth]{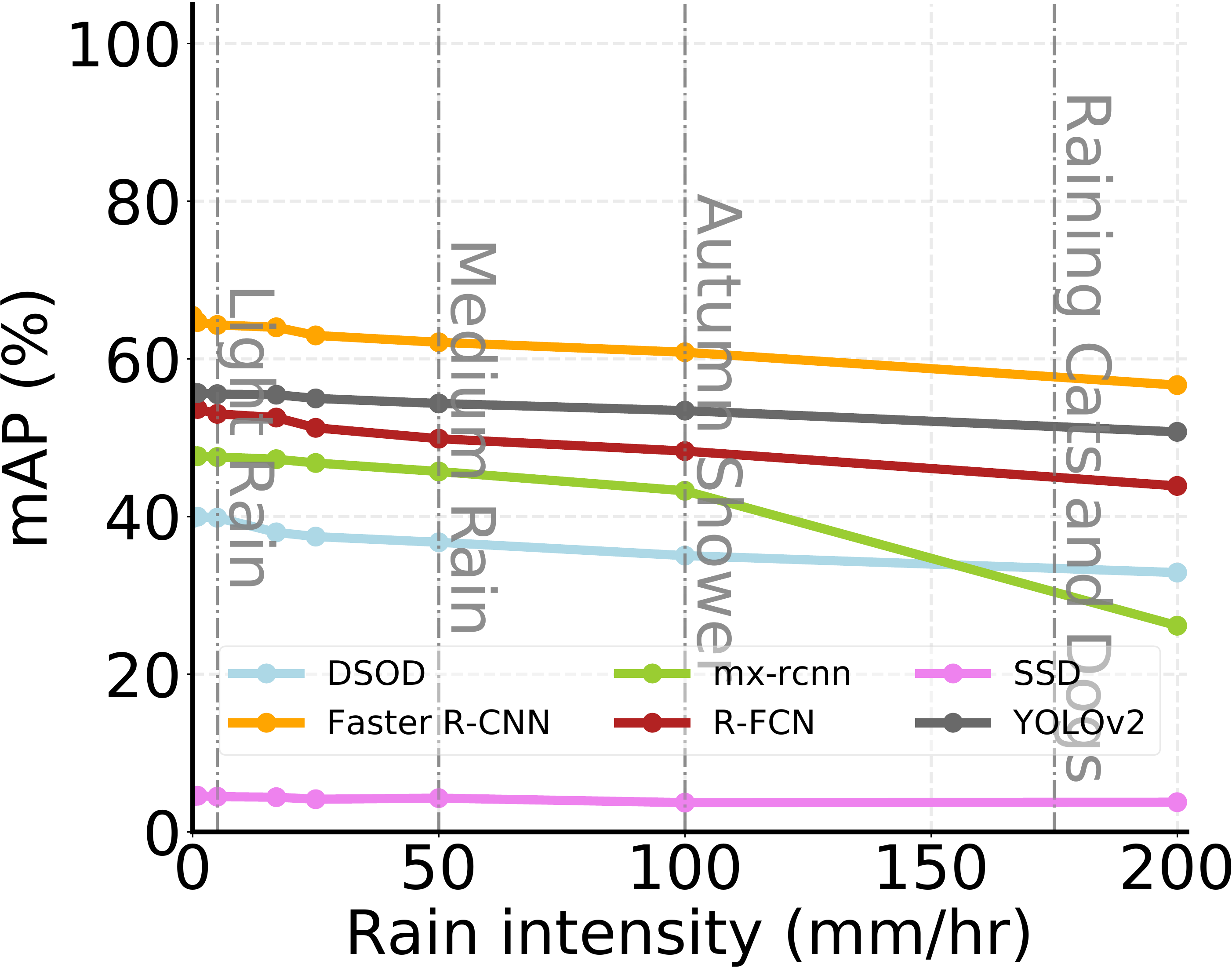}}\hspace{0.1\linewidth}
	\subfloat[Fog]{\includegraphics[width=0.40\linewidth]{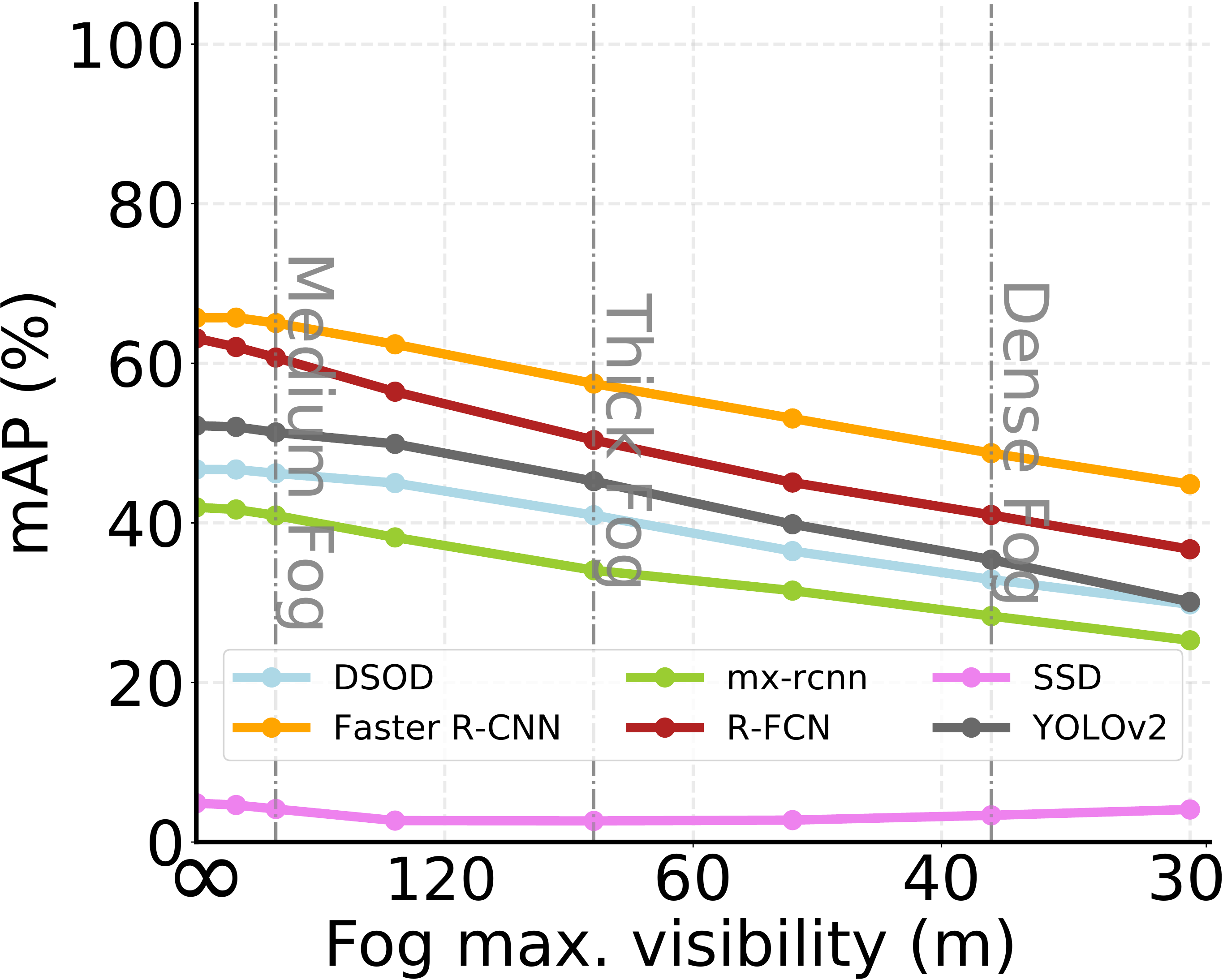}}
	\caption{Object detection performance on our weather augmented Kitti dataset as a function of rain fall rate (left) or fog visibility (right). The plots show the Coco mAP@[.1:.1:.9] accross car and pedestrians. While both fog and rain affect object detection, the effect of effect of fog is linear while rain is more chaotic.}
	\label{fig:obj_rain_fog_plot}
	\subfloat[Rain]{\includegraphics[width=0.40\linewidth]{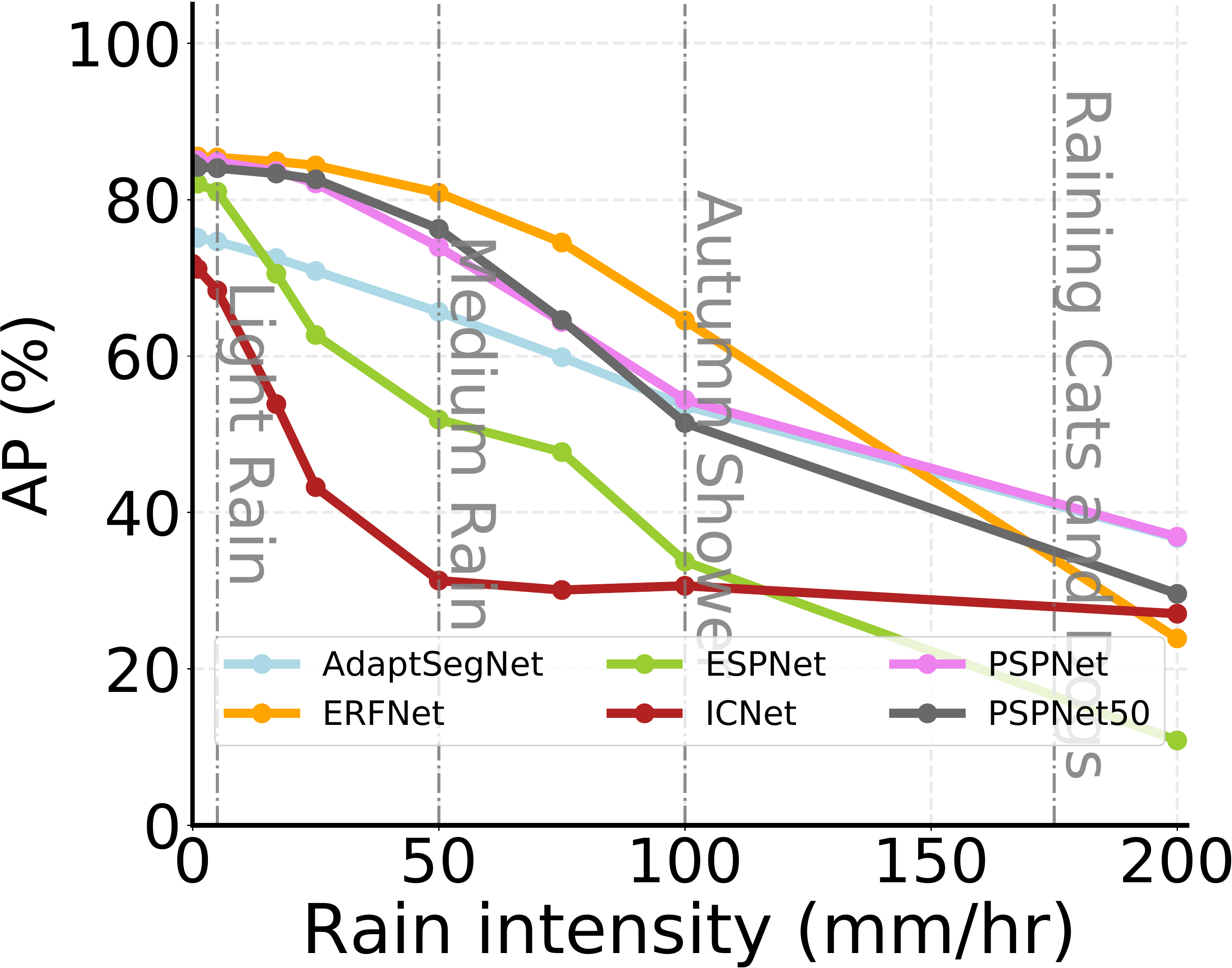}}\hspace{0.1\linewidth}
	\subfloat[Fog]{\includegraphics[width=0.40\linewidth]{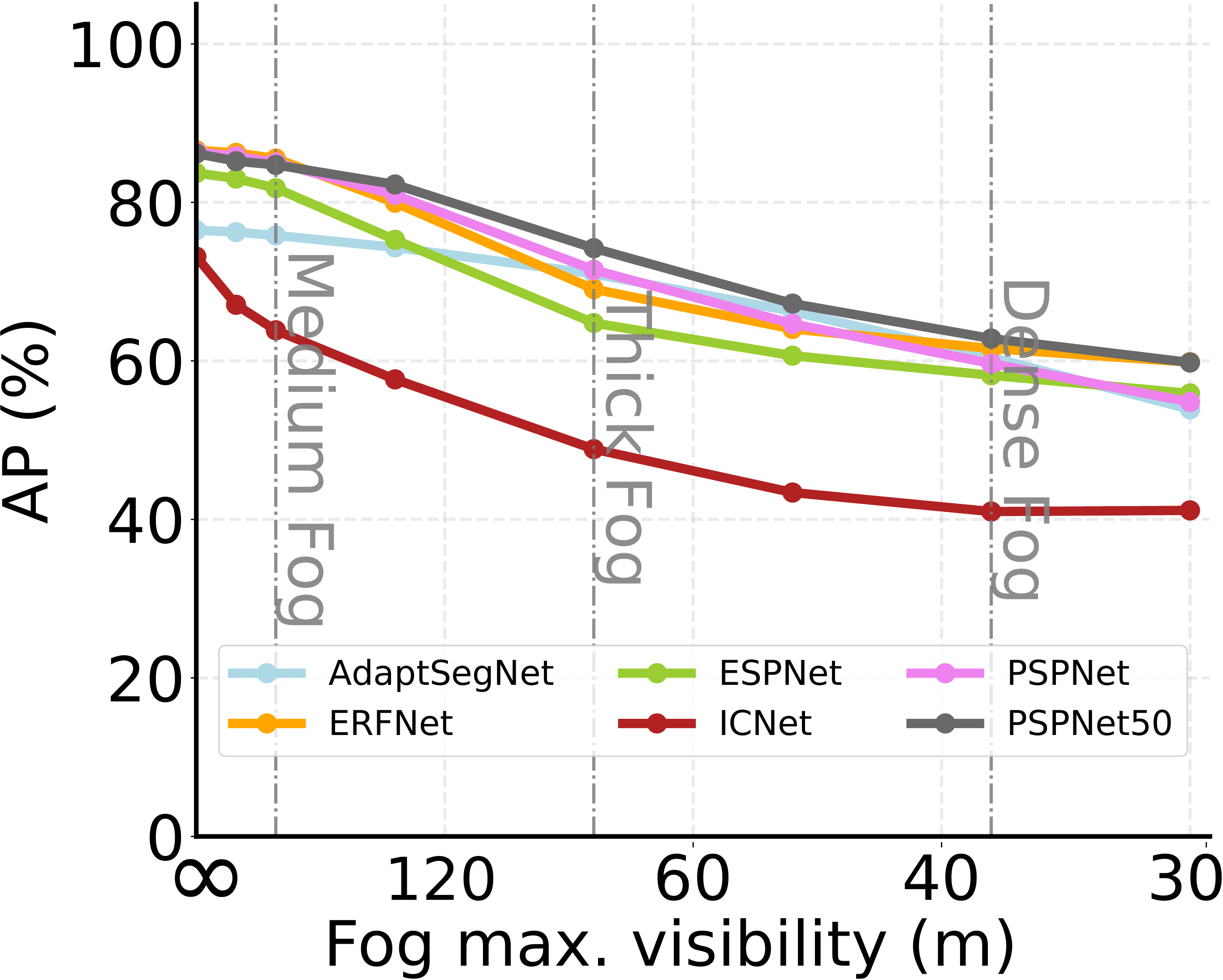}}
	\caption{Average Precision (AP) of the pixel-semantic prediction for clear and weather augmented Cityscapes dataset as a function of rain intensity (left) and fog extinction (right). This task is clearly more affected by the rain weather rather than the fog. This might be explained with the saliency patterns cause by rain streaks falling.}
	\label{fig:sem_rain_fog_plot}
\end{figure}

\begin{figure*}
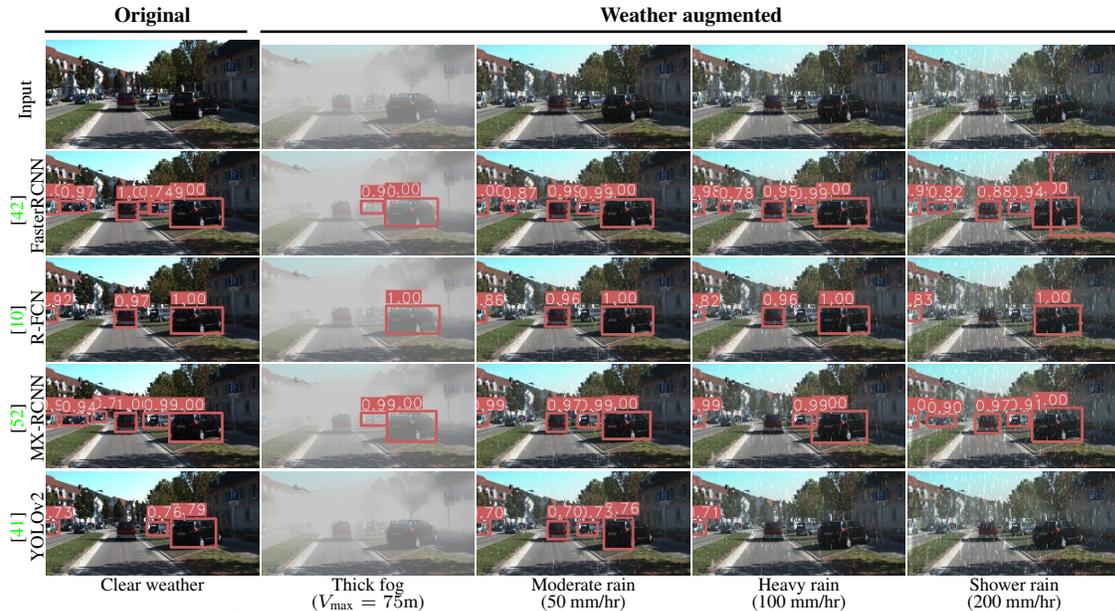

	\newcommand\weatheraugvizObjDetT[3]{\adjincludegraphics[width=0.162\linewidth,trim={{0.28346456692\width} 0 {0.12047244094\width} 0},clip]{figures/results_new/qualitative/objDetection/#2/#1#3_clear_tb.jpg}&\adjincludegraphics[width=0.162\linewidth,trim={{0.28346456692\width} 0 {0.12047244094\width} 0},clip]{figures/results_new/qualitative/objDetection/#2/#1#3_beta_80_tb.jpg}&\adjincludegraphics[width=0.162\linewidth,trim={{0.28346456692\width} 0 {0.12047244094\width} 0},clip]{figures/results_new/qualitative/objDetection/#2/#1#3_50mm_tb.jpg}&\adjincludegraphics[width=0.162\linewidth,trim={{0.28346456692\width} 0 {0.12047244094\width} 0},clip]{figures/results_new/qualitative/objDetection/#2/#1#3_100mm_tb.jpg}&\adjincludegraphics[width=0.162\linewidth,trim={{0.28346456692\width} 0 {0.12047244094\width} 0},clip]{figures/results_new/qualitative/objDetection/#2/#1#3_200mm_tb.jpg}}
	\newcommand\weatheraugvizObjDet[2]{\weatheraugvizObjDetT{#1}{#2}{_#2}}
	
	\newcommand{\objectdetectframe}{000095}
	\centering
	\scriptsize
	\setlength{\tabcolsep}{0.001\linewidth}
	\renewcommand{\arraystretch}{0.5}
	\begin{tabular}{cccccc}
		&\textbf{\footnotesize{Original}}&\multicolumn{4}{c}{\textbf{\footnotesize{Weather augmented}}}\\
		\cmidrule[1pt](r){2-2}\cmidrule[1pt](){3-6}
		\multirow{1}{*}[0.8cm]{\rotatebox{90}{Input}}&\weatheraugvizObjDetT{\objectdetectframe}{rendering}{}\\
 		\multirow{1}{*}[1.2cm]{\rotatebox{90}{~~~~~~\cite{ren2015faster}} \rotatebox{90}{FasterRCNN}}&\weatheraugvizObjDet{\objectdetectframe}{FasterRCNN}\\
		\multirow{1}{*}[0.85cm]{\rotatebox{90}{~~\cite{dai2016r}} \rotatebox{90}{R-FCN}}&\weatheraugvizObjDet{\objectdetectframe}{RFCN}\\
		\multirow{1}{*}[1.15cm]{\rotatebox{90}{~~~~~~\cite{yang2016exploit}} \rotatebox{90}{MX-RCNN}}&\weatheraugvizObjDet{\objectdetectframe}{MxRCNN}\\
		\multirow{1}{*}[1.0cm]{\rotatebox{90}{~~~~\cite{redmon2017yolo9000}} \rotatebox{90}{YOLOv2}}&\weatheraugvizObjDet{\objectdetectframe}{YoloV2}\\
		&Clear weather  & Thick fog 	& Moderate rain 	& Heavy rain 	& Shower rain \\
		&				& ($V_{\text{max}}=75$m) 	& (50~mm/hr) 	& (100~mm/hr) 		& (200~mm/hr) \\
	\end{tabular}
	
	\caption{Qualitative evaluation of object detection on our weather augmented Kitti dataset (cropped for visualization). From left to right, the original image (clear) and five weather augmented images.}
	\label{fig:eval_qualitative_objdet}
\end{figure*}

\begin{figure*}
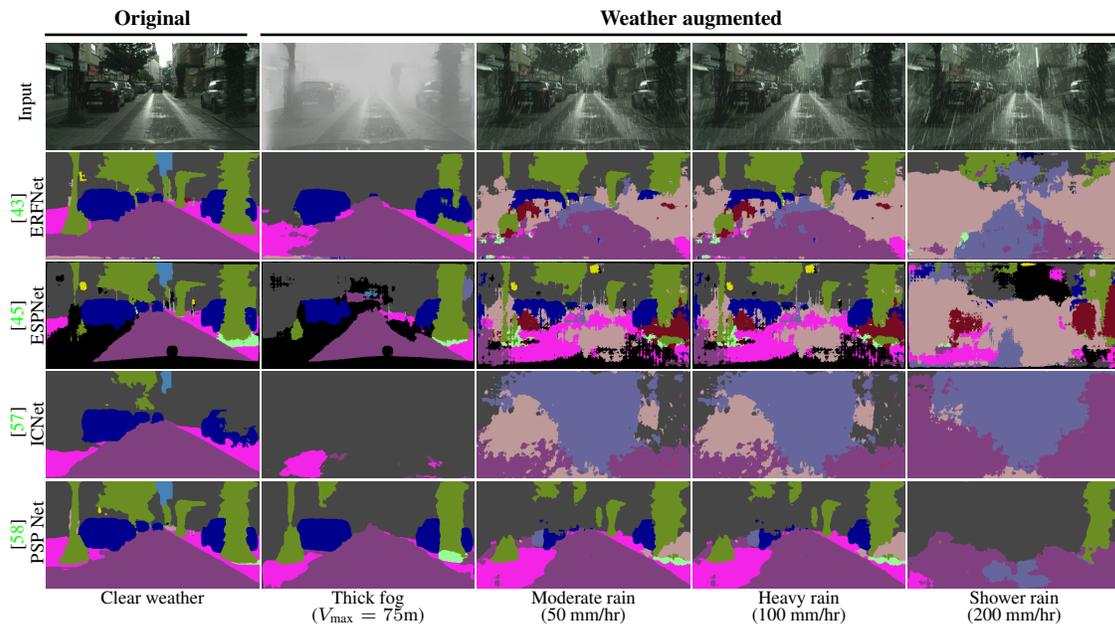

	\newcommand\weatheraugvizSegT[4]{\includegraphics[width=0.162\linewidth]{figures/results_new/qualitative/semanticSeg/#2/#1#3_clear_tb.#4}&\includegraphics[width=0.162\linewidth]{figures/results_new/qualitative/semanticSeg/#2/#1#3_beta_80_tb.#4}&\includegraphics[width=0.162\linewidth]{figures/results_new/qualitative/semanticSeg/#2/#1#3_100mm_tb.#4}&\includegraphics[width=0.162\linewidth]{figures/results_new/qualitative/semanticSeg/#2/#1#3_100mm_tb.#4}&\includegraphics[width=0.162\linewidth]{figures/results_new/qualitative/semanticSeg/#2/#1#3_200mm_tb.#4}}
	\newcommand\weatheraugvizSeg[2]{\weatheraugvizSegT{#1}{#2}{_#2}{png}}

	\centering
	\scriptsize
	\setlength{\tabcolsep}{0.001\linewidth}
	\renewcommand{\arraystretch}{0.5}
	\begin{tabular}{cccccc}
		&\textbf{\footnotesize{Original}}&\multicolumn{4}{c}{\textbf{\footnotesize{Weather augmented}}}\\
		\cmidrule[1pt](r){2-2}\cmidrule[1pt](){3-6}
		\multirow{1}{*}[0.8cm]{\rotatebox{90}{Input}}&\weatheraugvizSegT{bochum_000000_016260_leftImg8bit}{rendering}{}{jpg}\\
		
		\multirow{1}{*}[1.00cm]{\rotatebox{90}{~~~\cite{romera2018erfnet}} \rotatebox{90}{ERFNet}}&\weatheraugvizSeg{bochum_000000_016260_leftImg8bit}{ERFNet}\\
		\multirow{1}{*}[1.00cm]{\rotatebox{90}{~~~\cite{mehta2018espnet}} \rotatebox{90}{ESPNet}}&\weatheraugvizSeg{bochum_000000_016260_leftImg8bit}{ESPNet}\\
		\multirow{1}{*}[0.92cm]{\rotatebox{90}{~~\cite{ZhaoQSSJ17}} \rotatebox{90}{ICNet}}&\weatheraugvizSeg{bochum_000000_016260_leftImg8bit}{ICNet}\\
		\multirow{1}{*}[1.05cm]{\rotatebox{90}{~~~\cite{zhao2017pspnet}} \rotatebox{90}{PSP Net}}&\weatheraugvizSeg{bochum_000000_016260_leftImg8bit}{PSPNet}\\
		&Clear weather  & Thick fog 	& Moderate rain 	& Heavy rain 	& Shower rain \\
		&				& ($V_{\text{max}}=75$m) 	& (50~mm/hr) 	& (100~mm/hr) 		& (200~mm/hr) \\
	\end{tabular}
	
	\caption{Qualitative evaluation of semantic segmentation on weather augmented Cityscape dataset (cropped for visualization). From left to right, the original image (clear) and five weather augmented images.}
	\label{fig:eval_qualitative_seg}
\end{figure*}

\section{Studying the effects of rain}
\label{sec:evaluation}

In this section, we present an evaluation of the robustness of popular computer vision algorithms on Kitti~\cite{Geiger2012CVPR} and Cityscapes~\cite{Cordts2016Cityscapes} to demonstrate the usefulness of our rain rendering methodology. 
In doing so, we benefit from available ground truth labels to quantify the impact of rain and fog on two important tasks: object detection and semantic segmentation.
Both tasks are critical for outdoor computer vision systems such as mobile robotics. 
For a comprehensive study, we also render synthetic fog as in \cite{kahraman:hal-01620602,SDV18} which we describe in the supplementary. 
We use the maximum visibility distance $V_{\text{max}}$ to measure the fog intensity\footnote{The Koschmieder law defines the maximum visibility in fog as: $V_{\text{max}} = -\ln(C_T) / \beta$, where $\beta$ is the fog optical extinction, and $C_T = 0.05$ is the minimum identifiable contrast for humans~\cite{prokes2009atmospheric}. I.e., moderate/dense fog has $V_{\text{max}}$ of 375~m~($\beta=.008$) and 37.5~m~($\beta=.08$), respectively.}.
Unlike rain, fog is a steady weather that produces only a contrast attenuation function of the scene distance.

\subsection{Methodology}

We use the Kitti object benchmark~\cite{Geiger2012CVPR} (7480 images) for object detection and Cityscapes~\cite{Cordts2016Cityscapes} (2995 images) for segmentation, and evaluate 12 algorithms (6 per task) in 15 weather conditions.
The original (clear) serves as a baseline to which we compare the performance of 14 additional weather augmentation (7~types of rain, and 7~type of fog). 
For rain, we render from light rain to heavy storm corresponding to rates $R = \{0, 1, 5, 17, 25, 50, 100, 200\}$~mm/hr, and for fog, $V_{\text{max}} = \{\infty, 750, 375, 150, 75, 50, 40, 30\}$~m.
Kitti sequences are also used to demonstrate temporal performance in the supplementary video.

\subsubsection{Datasets preparation}
\label{sec:eval-datasets-prepare}

For realistic physical simulator (sec.~\ref{sec:raindrop-position}) and rainstreak photometric (sec.~\ref{sec:photometry-rainstreak}), intrinsic and extrinsic calibration are used to replicate the imaging sensor. 
For Kitti, we use sequence-wise or frame-wise calibration~\cite{Geiger2012CVPR,Geiger2013IJRR} with 6mm focal and 2ms exposure. As Cityscapes does not provide calibration, we use intrinsic from camera manufacturer with 5ms exposure and extrinsic is assumed similar to Kitti.

\noindent
Our method also requires the scene geometry (pixel depth) to  model accurately the light-particle interaction and the fog optical extinction. 
We estimate Kitti depth maps from RGB+Lidar with~\cite{jaritz2018sparse}, and Cityscapes depth from RGB only with MonoDepth~\cite{monodepth17}.
While perfect absolute depth is not required, a correct RGB-depth alignment is critical to avoid artifacts along geometrical edges.
Thus depths are post-processed with a guided-filter~\cite{barron2016fast} for better RGB-depth alignment. 

\subsubsection{Bad weather simulation}
We mimick the camera ego motion in the physical simulator to ensure realistic rainstreak orientation on still images and preserve temporal consistency in sequences.
Ego speed is extracted from GPS data when provided (Kitti sequences), or drawn uniformly in the $[0, 50]$~km/hr interval for Cityscapes semantics and in the $[0, 100]$~km/hr interval for Kitti object to reflect the urban and semi-urban scenarios, respectively.

\subsection{Object detection}
\label{sec:object-detection}

We evaluate the 15 augmented weathers on Kitti for 6~car/pedestrian pre-trained detection algorithms (with $\text{IoU}\geq.7$): DSOD~\cite{Shen2017DSOD}, Faster R-CNN~\cite{ren2015faster}, R-FCN~\cite{dai2016r}, SSD~\cite{liu2016ssd}, MX-RCNN~\cite{yang2016exploit}, and YOLOv2~\cite{redmon2017yolo9000}. Quantitative results for the Coco mAP@[.1:.1:.9] accross classes are shown in fig.~\ref{fig:obj_rain_fog_plot}. Relative to their clear-weather performance the 200~mm/hr rain is always at least 12\% worse and even drops to 25-30\% for R-FCN, SSD, and MX-RCNN, whereas Faster R-CNN and DSOD are the most robust to changes in fog and rain. Representative qualitative results are shown in fig.~\ref{fig:eval_qualitative_objdet} for 4 out of 6 algorithms to preserve space. We observe that, unlike fog, rain has a chaotic effect on object detection results whereas fog seems to affect the performance linearly with the depth. While all algorithms get affected by rain, when objects are large and facing the camera, most algorithms can still detect them.

\subsection{Semantic segmentation}

For semantic segmentation, the 15 weather augmented Cityscapes is evaluated for: AdaptSegNet~\cite{Tsai_adaptseg_2018}, ERFNet~\cite{romera2018erfnet}, ESPNet~\cite{mehta2018espnet}, ICNet~\cite{ZhaoQSSJ17}, PSPNet~\cite{zhao2017pspnet} and PSPNet(50)~\cite{zhao2017pspnet}.
Quantitative results are reported in fig.~\ref{fig:sem_rain_fog_plot} for both rain (a) and fog (b). As opposed to object detection algorithms which demonstrated significant robustness to moderately high rainfall rates, here the algorithms seem to breakdown in similar conditions. Indeed, all techniques see their performance drop by a minimum of 30\% under heavy fog, and almost 60\% under strong rain. Interestingly, some curves cross, which indicates that different algorithms behave different under rain. ESPNet for example, ranks among the top 3 in clear weather but drops relatively by a staggering 85\% and ranks last in when it is raining cats and dogs (200~mm/hr). Corresponding qualitative results are shown in fig.~\ref{fig:eval_qualitative_seg} for 4 out of 6 algorithms to preserve space. Although the effect of rain may appear minimal visually, it greatly affects the output of all segmentation algorithms evaluated.

\section{Improving robustness to rain}
Using our rain-rendered database, we now demonstrate its usefulness for improving robustness to rain through extensive evaluations on synthetic and real rain databases.

\subsection{Methodology}
We select Faster-RCNN \cite{ren2015faster} for object detection and PSPNet~\cite{zhao2017pspnet} for semantic segmentation due to their public train implementation and good performance.
While the ultimate goal is to improve robustness to rain, we aim at learning a model showing robustness to both clear weather and large variety of rains.
Because rain significantly alters the appearance of the scene, we found that training from scratch with rain fails to converge.
Instead, we refine our \textit{untuned} models using curriculum learning~\cite{bengio2009curriculum} on rain intensity in ascending order (25, then 50, then 75 and finally 100mm/hr rain). The final model is referred as \textit{finetuned} and is evaluated against various weather conditions.

Each of the 4 refinement passes uses 1000 images of the corresponding rain fallrate, and trains for 10 epochs with 0.0004 learning rate and 0.9 momentum.

\begin{figure}
	\centering
	\subfloat[Object detection mAP (\%)]{\includegraphics[width=0.41\linewidth]{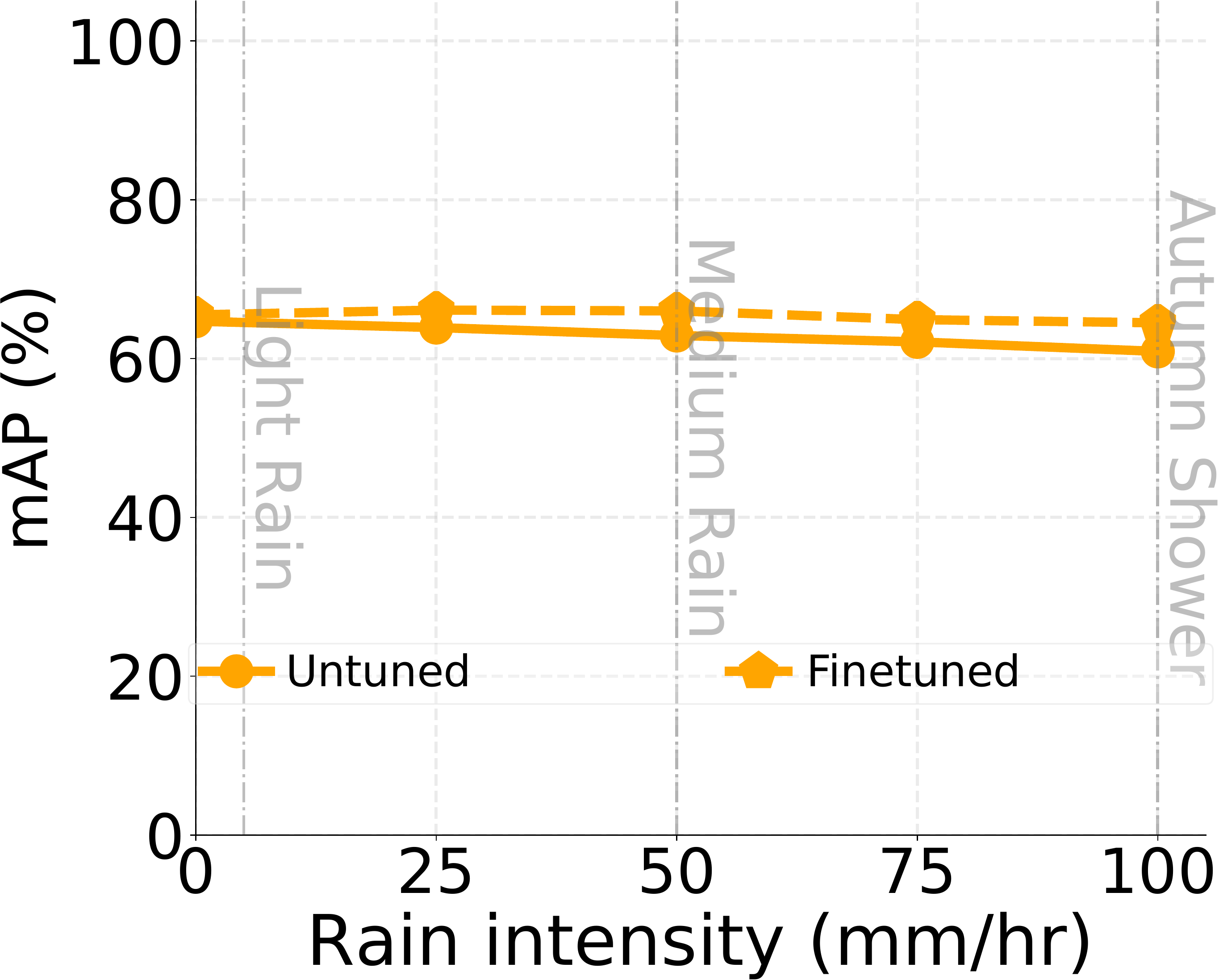}\hspace{.06\linewidth}}\hspace{.05\linewidth}
	\subfloat[Semantic segmentation AP (\%)]{\includegraphics[width=0.41\linewidth]{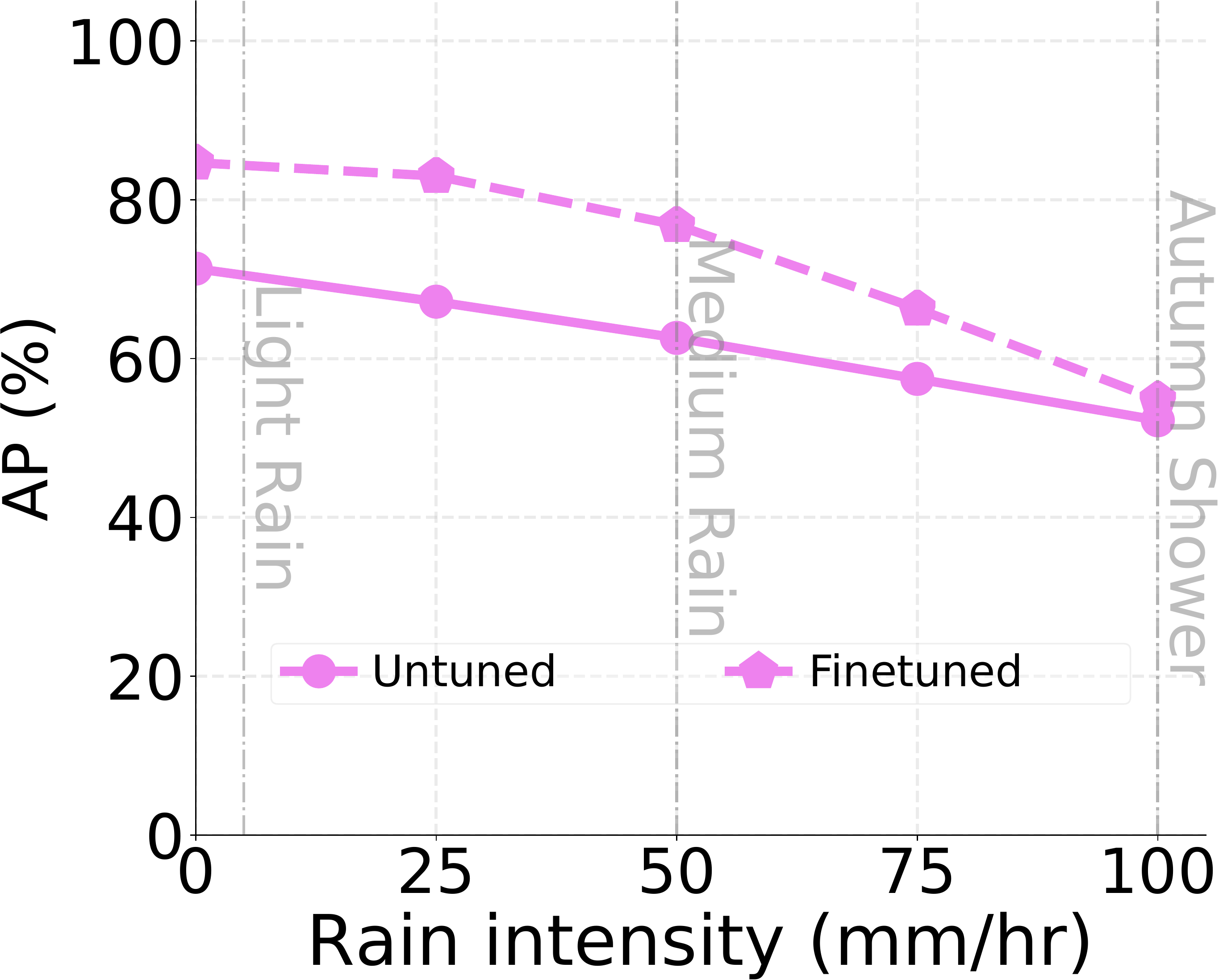}\hspace{.06\linewidth}}
	\caption{Performance on synthetic data for object detection (Faster-RCNN~\cite{ren2015faster}) and semantic segmentation (PSPNet~\cite{zhao2017pspnet}) when finetuned with our rendering pipeline. Both finetuned models show increased robustness to rain events and clear weather.}
	\label{fig:impr_finetune}
	\centering
	\subfloat[{Object detection mAP (\%)}]{
		\scriptsize
		\setlength{\tabcolsep}{0.18cm}
		\renewcommand{\arraystretch}{0.5}
		\begin{tabular}{lcc}
			& Clear & Rainy \\
			\midrule
			Untuned & 19.5 & 10.1 \\
			\midrule
			Finetuned \tiny{(ours)} & \textbf{20.1} & \textbf{11.6}\\
			\bottomrule
	\end{tabular}}\hspace{.05\linewidth}
	\subfloat[{Semantic segmentation AP (\%)}]{
		\scriptsize
		\setlength{\tabcolsep}{0.18cm}
		\renewcommand{\arraystretch}{0.5}
		\begin{tabular}{lcc}
			& Clear & Rainy \\
			\midrule
			Untuned & \textbf{40.8} & 18.7 \\
			\midrule
			Finetuned \tiny{(ours)} & 39.0 & \textbf{25.6}\\
			\bottomrule
	\end{tabular}}
	\vspace{.5em}
	\caption{Real rain performance on nuScenes \cite{caesar2019nuscenes} datasets for object detection (Faster-RCNN~\cite{ren2015faster}) and semantic segmentation (PSPNet~\cite{zhao2017pspnet}).}
	\label{fig:nuscenes}
\end{figure}

\subsection{Synthetic performance}
The synthetic evaluation is conducted on our augmented databases using 1000 versatile unseen images, with either no-rain (clear) or rain up to 200mm/hr. 
Fig.~\ref{fig:impr_finetune} shows the performance of our \textit{untuned} and \textit{finetuned} model for object detection (Faster-RCNN \cite{ren2015faster}), and semantic segmentation (PSPNet \cite{zhao2017pspnet}).
We observe an obvious improvement in both tasks and additional increase in robustness even in the clear weather when refined using our augmented rain.
The intuition here is that when facing adverse weather, the network learns to focus on relevant features for both tasks and thus gain robustness.
For Faster-RCNN, the finetuned detection performance is nearly constant in the 0-100mm/hr. Explicitly, it drops to $64.5\%$ whereas the untuned model drops to $60.9\%$. We also tested on stronger (unseen) 200mm/hr fallrate and our finetuned Faster-RCNN got an mAP of $62.4\%$ (versus $55.4\%$ when untuned).
For PSPNet, the segmentation exhibits a significative improvement when refined although at 100mm/hr the model is not fully able to compensate the effect of rain and drops to $54.0\%$ versus $52.0\%$ when untuned.

\subsection{Real rain performance}
We use the recent NuScenes dataset~\cite{caesar2019nuscenes} which provides coarse weather meta-data and evaluate object detection and segmentation when \textit{untuned} or \textit{finetuned}.
Since meteorological stations only report hour average, precipitation in mm/hr cannot be retrieved frame-wise so we cluster frames into clear and rainy.
For objects, we study the mAP on 2000 nuScenes images (1000 clear + 1000 rainy). For segmentation, since semantic labels are not provided we evaluate AP on 50 images (25+25) which we carefully annotated ourselves.
There is here a large domain gap between training (Kitti/Cityscapes with our synthetic rain) and evaluation (nuScenes with real rain). 
Still, Fig.~\ref{fig:nuscenes} shows that ours \textit{finetuned} leads to performance increase in all real rainy scenes for both object (+14.9\%) and semantic tasks (+36.6\%). 
In clear weather ours \textit{finetuned} performs on par with the \textit{untuned} version. This demonstrates the usefulness of our physics-based rain rendering for real rain scenarios. 

More comparative and qualitative results are present in the supplementary.
\section{Discussion}

In this paper, we presented the first intensity-controlled physical framework for augmenting existing image databases with realistic rain.

\textbf{Limitations.} While we demonstrated highly realistic rain rendering results, our approach still has limitations that set the stage for future work. The main limitation is the approximation of the lighting conditions (sec.~\ref{sec:rendering.rain}). 
While it was empirically determined our environment map yielded reasonable results (fig.~\ref{fig:pano_ours_comparison}), it may under/over estimate the scene radiance when the sky is not/too visible.
This approximation is more visible when streaks are imaged against a darker sky. %

Second, we make explicit distinction between fog-like rain and drops imaged on more than 1 pixel, individually rendered as streaks. 
While this distinction is widely used in the literature \cite{garg2007vision,garg2005does,de2012fast,Li_2018_ECCV} it causes an inappropriate sharp distinction between fog-like and streaks.
A possible solution would be to render \emph{all} drops as streaks weighting them as a function of their imaging surface. 
However, our experiments show it comes at a prohibitive computation cost.

Finally, our rendering framework does not model wet surface and the splashes rain makes on surfaces~\cite{garg2007material}. 
Properly modeling these effects would require richer illumination and 3D information about the scene, including the location and nature of all surfaces. 
Nevertheless, it is likely that as scene geometry estimation techniques progress~\cite{monodepth17}, rendering more and more of these effects will be achievable. 

\textbf{Usage by the community.}
Our framework is readily usable to augment existing image with realistic rainy conditions.
The weather augmented Kitti and Cityscapes, as well as the code to generate rain and fog in arbitrary images/sequences is available on the project page.

\section*{Acknowledgements}

This work was partially supported by the \textit{Service de coopération et d'action culturelle du Consulat général de France à Québec} with the grant Samuel de Champlain. We gratefully thank Pierre Bourré for his priceless technical help, Aitor Gomez Torres for his initial input, and Srinivas Narasimhan for letting us reuse the physical simulator.

{\small
\bibliographystyle{ieee_fullname}
\bibliography{egbib}

\begin{thebibliography}{10}\itemsep=-1pt

\bibitem{atlas1973doppler}
David Atlas, RC Srivastava, and Rajinder~S Sekhon.
\newblock Doppler radar characteristics of precipitation at vertical incidence.
\newblock {\em Reviews of Geophysics}, 11(1):1--35, 1973.

\bibitem{barnum2010analysis}
Peter~C Barnum, Srinivasa Narasimhan, and Takeo Kanade.
\newblock Analysis of rain and snow in frequency space.
\newblock {\em International Journal of Computer Vision}, 86(2-3):256, 2010.

\bibitem{barron2016fast}
Jonathan~T Barron and Ben Poole.
\newblock The fast bilateral solver.
\newblock In {\em European Conference on Computer Vision}, 2016.

\bibitem{bengio2009curriculum}
Yoshua Bengio, J{\'e}r{\^o}me Louradour, Ronan Collobert, and Jason Weston.
\newblock Curriculum learning.
\newblock In {\em International Conference on Machine Learning}, pages 41--48.
  ACM, 2009.

\bibitem{caesar2019nuscenes}
Holger Caesar and et al.
\newblock nuscenes: A multimodal dataset for autonomous driving.
\newblock {\em preprint arXiv:1903.11027}, 2019.

\bibitem{Cameron2005}
Christopher Cameron.
\newblock Hallucinating environment maps from single images.
\newblock Technical report, 2005.

\bibitem{chen2013generalized}
Yi-Lei Chen and Chiou-Ting Hsu.
\newblock A generalized low-rank appearance model for spatio-temporally
  correlated rain streaks.
\newblock In {\em IEEE International Conference on Computer Vision}, pages
  1968--1975, 2013.

\bibitem{Cordts2016Cityscapes}
Marius Cordts, Mohamed Omran, Sebastian Ramos, Timo Rehfeld, Markus Enzweiler,
  Rodrigo Benenson, Uwe Franke, Stefan Roth, and Bernt Schiele.
\newblock The cityscapes dataset for semantic urban scene understanding.
\newblock In {\em IEEE Conference on Computer Vision and Pattern Recognition},
  2016.

\bibitem{creus2013r4}
Carles Creus and Gustavo~A Patow.
\newblock R4: Realistic rain rendering in realtime.
\newblock {\em Computers \& Graphics}, 37(1-2):33--40, 2013.

\bibitem{dai2016r}
Jifeng Dai, Yi Li, Kaiming He, and Jian Sun.
\newblock R-fcn: Object detection via region-based fully convolutional
  networks.
\newblock In {\em Advances in Neural Information Processing Systems}, pages
  379--387, 2016.

\bibitem{de2012fast}
Raoul de Charette, Robert Tamburo, Peter~C Barnum, Anthony Rowe, Takeo Kanade,
  and Srinivasa~G Narasimhan.
\newblock Fast reactive control for illumination through rain and snow.
\newblock In {\em International Conference on Computational Photography}, 2012.

\bibitem{eigen2013restoring}
David Eigen, Dilip Krishnan, and Rob Fergus.
\newblock Restoring an image taken through a window covered with dirt or rain.
\newblock In {\em IEEE International Conference on Computer Vision}, pages
  633--640, 2013.

\bibitem{garg2007material}
Kshitiz Garg, Gurunandan Krishnan, and Shree~K Nayar.
\newblock Material based splashing of water drops.
\newblock In {\em Eurographics Conference on Rendering Techniques}, 2007.

\bibitem{garg2004detection}
Kshitiz Garg and Shree~K Nayar.
\newblock Detection and removal of rain from videos.
\newblock In {\em IEEE Conference on Computer Vision and Pattern Recognition},
  2004.

\bibitem{garg2005does}
Kshitiz Garg and Shree~K Nayar.
\newblock When does a camera see rain?
\newblock In {\em IEEE International Conference on Computer Vision}, volume~2,
  pages 1067--1074. IEEE, 2005.

\bibitem{garg2006photorealistic}
Kshitiz Garg and Shree~K Nayar.
\newblock Photorealistic rendering of rain streaks.
\newblock In {\em ACM Transactions on Graphics (SIGGRAPH)}, volume~25, pages
  996--1002. ACM, 2006.

\bibitem{garg2007vision}
Kshitiz Garg and Shree~K Nayar.
\newblock Vision and rain.
\newblock {\em International Journal of Computer Vision}, 75(1):3--27, 2007.

\bibitem{gatys2016image}
Leon~A Gatys, Alexander~S Ecker, and Matthias Bethge.
\newblock Image style transfer using convolutional neural networks.
\newblock In {\em IEEE Conference on Computer Vision and Pattern Recognition},
  2016.

\bibitem{Geiger2013IJRR}
Andreas Geiger, Philip Lenz, Christoph Stiller, and Raquel Urtasun.
\newblock Vision meets robotics: The kitti dataset.
\newblock {\em International Journal of Robotics Research}, 2013.

\bibitem{Geiger2012CVPR}
Andreas Geiger, Philip Lenz, and Raquel Urtasun.
\newblock Are we ready for autonomous driving? {T}he {KITTI} vision benchmark
  suite.
\newblock In {\em IEEE Conference on Computer Vision and Pattern Recognition},
  2012.

\bibitem{monodepth17}
Cl{\'{e}}ment Godard, Oisin {Mac Aodha}, and Gabriel~J. Brostow.
\newblock Unsupervised monocular depth estimation with left-right consistency.
\newblock In {\em IEEE Conference on Computer Vision and Pattern Recognition},
  2017.

\bibitem{halimeh2009raindrop}
Jad~C Halimeh and Martin Roser.
\newblock Raindrop detection on car windshields using geometric-photometric
  environment construction and intensity-based correlation.
\newblock In {\em IEEE Intelligent Vehicles Symposium}, pages 610--615. IEEE,
  2009.

\bibitem{holdgeoffroy-cvpr-19}
Yannick Hold-Geoffroy, Akshaya Athawale, and Jean-Fran\c{c}ois Lalonde.
\newblock Deep sky modeling for single image outdoor lighting estimation.
\newblock In {\em IEEE Conference on Computer Vision and Pattern Recognition},
  2019.

\bibitem{horn1986robot}
Berthold Horn, Berthold Klaus, and Paul Horn.
\newblock {\em Robot vision}.
\newblock MIT press, 1986.

\bibitem{jacobs07amos}
Nathan Jacobs, Nathaniel Roman, and Robert Pless.
\newblock Consistent temporal variations in many outdoor scenes.
\newblock In {\em IEEE Conference on Computer Vision and Pattern Recognition},
  2007.

\bibitem{jaritz2018sparse}
Maximilian Jaritz, Raoul de Charette, Emilie Wirbel, Xavier Perrotton, and
  Fawzi Nashashibi.
\newblock Sparse and dense data with {CNNs}: Depth completion and semantic
  segmentation.
\newblock In {\em International Conference on 3D Vision}, 2018.

\bibitem{johnson2016perceptual}
Justin Johnson, Alexandre Alahi, and Li Fei-Fei.
\newblock Perceptual losses for real-time style transfer and super-resolution.
\newblock In {\em European Conference on Computer Vision}, 2016.

\bibitem{johnson2016driving}
Matthew Johnson-Roberson, Charles Barto, Rounak Mehta, Sharath~Nittur Sridhar,
  Karl Rosaen, and Ram Vasudevan.
\newblock Driving in the matrix: Can virtual worlds replace human-generated
  annotations for real world tasks?
\newblock In {\em International Conference on Robotics and Automation}, 2016.

\bibitem{kahraman:hal-01620602}
Sule Kahraman and Raoul de Charette.
\newblock {Influence of Fog on Computer Vision Algorithms}.
\newblock Research report, {Inria Paris}, Sept. 2017.

\bibitem{Laffont14}
Pierre-Yves Laffont, Zhile Ren, Xiaofeng Tao, Chao Qian, and James Hays.
\newblock Transient attributes for high-level understanding and editing of
  outdoor scenes.
\newblock {\em ACM Transactions on Graphics (SIGGRAPH)}, 33(4), 2014.

\bibitem{lalonde2009webcam}
Jean-Fran{\c{c}}ois Lalonde, Alexei~A Efros, and Srinivasa~G Narasimhan.
\newblock Webcam clip art: Appearance and illuminant transfer from time-lapse
  sequences.
\newblock In {\em ACM Transactions on Graphics (SIGGRAPH Asia)}, volume~28,
  page 131, 2009.

\bibitem{Li_2018_ECCV}
Ruoteng Li, Robby~T. Tan, and Loong-Fah Cheong.
\newblock Robust optical flow in rainy scenes.
\newblock In {\em European Conference on Computer Vision}, September 2018.

\bibitem{li2016rain}
Yu Li, Robby~T Tan, Xiaojie Guo, Jiangbo Lu, and Michael~S Brown.
\newblock Rain streak removal using layer priors.
\newblock In {\em IEEE Conference on Computer Vision and Pattern Recognition},
  pages 2736--2744, 2016.

\bibitem{liu2016ssd}
Wei Liu, Dragomir Anguelov, Dumitru Erhan, Christian Szegedy, Scott Reed,
  Cheng-Yang Fu, and Alexander~C Berg.
\newblock Ssd: Single shot multibox detector.
\newblock In {\em European Conference on Computer Vision}, 2016.

\bibitem{luo2015removing}
Yu Luo, Yong Xu, and Hui Ji.
\newblock Removing rain from a single image via discriminative sparse coding.
\newblock In {\em IEEE International Conference on Computer Vision}, pages
  3397--3405, 2015.

\bibitem{RobotCarDatasetIJRR}
Will Maddern, Geoff Pascoe, Chris Linegar, and Paul Newman.
\newblock {1 Year, 1000km: The Oxford RobotCar Dataset}.
\newblock {\em International Journal of Robotics Research}, 36(1):3--15, 2017.

\bibitem{marshall1948distribution}
John~S Marshall and W~Mc~K Palmer.
\newblock The distribution of raindrops with size.
\newblock {\em Journal of meteorology}, 5(4):165--166, 1948.

\bibitem{narasimhan2002all}
Srinivasa~G Narasimhan, Chi Wang, and Shree~K Nayar.
\newblock All the images of an outdoor scene.
\newblock In {\em European Conference on Computer Vision}, 2002.

\bibitem{potmesil1981lens}
Michael Potmesil and Indranil Chakravarty.
\newblock A lens and aperture camera model for synthetic image generation.
\newblock {\em ACM SIGGRAPH Computer Graphics}, 15(3):297--305, 1981.

\bibitem{prokes2009atmospheric}
Ales Prokes.
\newblock Atmospheric effects on availability of free space optics systems.
\newblock {\em Optical Engineering}, 48(6):066001, 2009.

\bibitem{redmon2017yolo9000}
Joseph Redmon and Ali Farhadi.
\newblock Yolo9000: better, faster, stronger.
\newblock In {\em IEEE Conference on Computer Vision and Pattern Recognition},
  2017.

\bibitem{ren2015faster}
Shaoqing Ren, Kaiming He, Ross Girshick, and Jian Sun.
\newblock Faster r-cnn: Towards real-time object detection with region proposal
  networks.
\newblock In {\em Advances in Neural Information Processing Systems}, pages
  91--99, 2015.

\bibitem{romera2018erfnet}
Eduardo Romera, Jos{\'e}~M Alvarez, Luis~M Bergasa, and Roberto Arroyo.
\newblock Erfnet: Efficient residual factorized convnet for real-time semantic
  segmentation.
\newblock {\em IEEE Transactions on Intelligent Transportation Systems},
  19(1):263--272, 2018.

\bibitem{rousseau2006realistic}
Pierre Rousseau, Vincent Jolivet, and Djamchid Ghazanfarpour.
\newblock Realistic real-time rain rendering.
\newblock {\em Computers \& Graphics}, 30(4):507--518, 2006.

\bibitem{mehta2018espnet}
Mohammad~Rastegari Sachin~Mehta, Anat Caspi, Linda Shapiro, and Hannaneh
  Hajishirzi.
\newblock Espnet: Efficient spatial pyramid of dilated convolutions for
  semantic segmentation.
\newblock In {\em European Conference on Computer Vision}, 2018.

\bibitem{SDV18}
Christos Sakaridis, Dengxin Dai, and Luc Van~Gool.
\newblock Semantic foggy scene understanding with synthetic data.
\newblock {\em International Journal of Computer Vision}, 126(9):973--992, Sep
  2018.

\bibitem{Shen2017DSOD}
Zhiqiang Shen, Zhuang Liu, Jianguo Li, Yu-Gang Jiang, Yurong Chen, and
  Xiangyang Xue.
\newblock Dsod: Learning deeply supervised object detectors from scratch.
\newblock In {\em IEEE International Conference on Computer Vision}, 2017.

\bibitem{tatarchuk2006artist}
Natalya Tatarchuk.
\newblock Artist-directable real-time rain rendering in city environments.
\newblock In {\em ACM SIGGRAPH Courses}, pages 23--64. ACM, 2006.

\bibitem{Tsai_adaptseg_2018}
Yi-Hsuan Tsai, Wei-Chih Hung, Samuel Schulter, Kihyuk Sohn, Ming-Hsuan Yang,
  and Manmohan Chandraker.
\newblock Learning to adapt structured output space for semantic segmentation.
\newblock In {\em IEEE Conference on Computer Vision and Pattern Recognition},
  2018.

\bibitem{van1997numerical}
John~H van Boxel et~al.
\newblock Numerical model for the fall speed of rain drops in a rain fall
  simulator.
\newblock In {\em Workshop on wind and water erosion}, pages 77--85, 1997.

\bibitem{weber2015multiscale}
Yoann Weber, Vincent Jolivet, Guillaume Gilet, and Djamchid Ghazanfarpour.
\newblock A multiscale model for rain rendering in real-time.
\newblock {\em Computers \& Graphics}, 50:61--70, 2015.

\bibitem{yang2016exploit}
Fan Yang, Wongun Choi, and Yuanqing Lin.
\newblock Exploit all the layers: Fast and accurate cnn object detector with
  scale dependent pooling and cascaded rejection classifiers.
\newblock In {\em IEEE Conference on Computer Vision and Pattern Recognition},
  2016.

\bibitem{yang2017deep}
Wenhan Yang, Robby~T Tan, Jiashi Feng, Jiaying Liu, Zongming Guo, and Shuicheng
  Yan.
\newblock Deep joint rain detection and removal from a single image.
\newblock In {\em IEEE Conference on Computer Vision and Pattern Recognition},
  pages 1357--1366, 2017.

\bibitem{yu2018bdd100k}
Fisher Yu, Wenqi Xian, Yingying Chen, Fangchen Liu, Mike Liao, Vashisht
  Madhavan, and Trevor Darrell.
\newblock Bdd100k: A diverse driving video database with scalable annotation
  tooling.
\newblock {\em arXiv preprint arXiv:1805.04687}, 2018.

\bibitem{zhang2018density}
He Zhang and Vishal~M Patel.
\newblock Density-aware single image de-raining using a multi-stream dense
  network.
\newblock In {\em IEEE Conference on Computer Vision and Pattern Recognition},
  pages 695--704, 2018.

\bibitem{zhang2017image}
He Zhang, Vishwanath Sindagi, and Vishal~M Patel.
\newblock Image de-raining using a conditional generative adversarial network.
\newblock {\em arXiv preprint arXiv:1701.05957}, 2017.

\bibitem{ZhaoQSSJ17}
Hengshuang Zhao, Xiaojuan Qi, Xiaoyong Shen, Jianping Shi, and Jiaya Jia.
\newblock Icnet for real-time semantic segmentation on high-resolution images.
\newblock In {\em European Conference on Computer Vision}, 2017.

\bibitem{zhao2017pspnet}
Hengshuang Zhao, Jianping Shi, Xiaojuan Qi, Xiaogang Wang, and Jiaya Jia.
\newblock Pyramid scene parsing network.
\newblock In {\em IEEE Conference on Computer Vision and Pattern Recognition},
  2017.

\bibitem{zhu2017unpaired}
Jun-Yan Zhu, Taesung Park, Phillip Isola, and Alexei~A Efros.
\newblock Unpaired image-to-image translation using cycle-consistent
  adversarial networks.
\newblock 2017.

\end{thebibliography}
}

\end{document}